%% file: AGQA.tex
\documentclass[final]{cvpr}

\usepackage{times}
\usepackage{epsfig}
\usepackage{graphicx}
\usepackage{amsmath}
\usepackage{amssymb}
\usepackage{multirow}
\usepackage{longtable}
\usepackage{lipsum}
\usepackage[table,xcdraw]{xcolor}
\usepackage[ruled,vlined]{algorithm2e}
\usepackage{rotating}
\usepackage{tabularx}

\definecolor{objectColor}{rgb}{0.6, 0, 0}
\newcommand{\object}[1]{{\color{objectColor}{#1}}}
\definecolor{relationshipColor}{rgb}{0.415, 0.658, 0.309}
\newcommand{\relationship}[1]{{\color{relationshipColor}{#1}}}
\definecolor{actionColor}{rgb}{0.901, 0.568, 0.219}
\newcommand{\action}[1]{{\color{actionColor}{#1}}}
\definecolor{timeColor}{rgb}{0.403, 0.305, 0.654}
\newcommand{\temporal}[1]{{\color{timeColor}{#1}}}

\usepackage[pagebackref=true,breaklinks=true,colorlinks,bookmarks=false]{hyperref}

\def\cvprPaperID{2764} 
\def\confYear{CVPR 2021}

\begin{document}

\title{AGQA: A Benchmark for Compositional Spatio-Temporal Reasoning}

\author{Madeleine Grunde-McLaughlin\\
University of Pennsylvania\\
{\tt\small mgrund@sas.upenn.edu}

\and
Ranjay Krishna\\
Stanford University\\
{\tt\small ranjaykrishna@cs.stanford.edu}

\and
Maneesh Agrawala\\
Stanford University\\
{\tt\small maneesh@cs.stanford.edu}

}
\maketitle

\begin{abstract}
\input{sections/00_abstract}
\end{abstract}

\input{sections/01_introduction}

\input{sections/02_related}

\input{sections/03_method}

\input{sections/04_experiments}

\input{sections/05_discussions}
{\small
\bibliographystyle{ieee_fullname}
\bibliography{references}
}

\newpage

\input{sections/06_supplementary}

\end{document}

%% file: sections/00_abstract.tex
Visual events are a composition of temporal actions involving actors spatially interacting with objects.
When developing computer vision models that can reason about compositional spatio-temporal events, we need benchmarks that can analyze progress and uncover shortcomings.
Existing video question answering benchmarks are useful, but they often conflate multiple sources of error into one accuracy metric and have strong biases that models can exploit, making it difficult to pinpoint model weaknesses.
We present Action Genome Question Answering (AGQA), a new benchmark for compositional spatio-temporal reasoning. AGQA contains $192M$ unbalanced question answer pairs for $9.6K$ videos. We also provide a 
balanced subset of $3.9M$ question answer pairs, $3$ orders of magnitude larger than existing benchmarks, that minimizes bias by balancing the answer distributions and types of question structures. Although human evaluators marked $86.02\%$ of our question-answer pairs as correct, the best model achieves only $47.74\%$ accuracy. 
In addition, AGQA introduces multiple training/test splits to test for various reasoning abilities, including generalization to novel compositions, to indirect references, and to more compositional steps. 
Using AGQA, we evaluate modern visual reasoning systems, demonstrating that the best models barely perform better than non-visual baselines exploiting linguistic biases and that none of the existing models generalize to novel compositions unseen during training.

%% file: sections/01_introduction.tex
\section{Introduction}

\begin{figure}[t]
    \centering
    \includegraphics[width=\columnwidth]{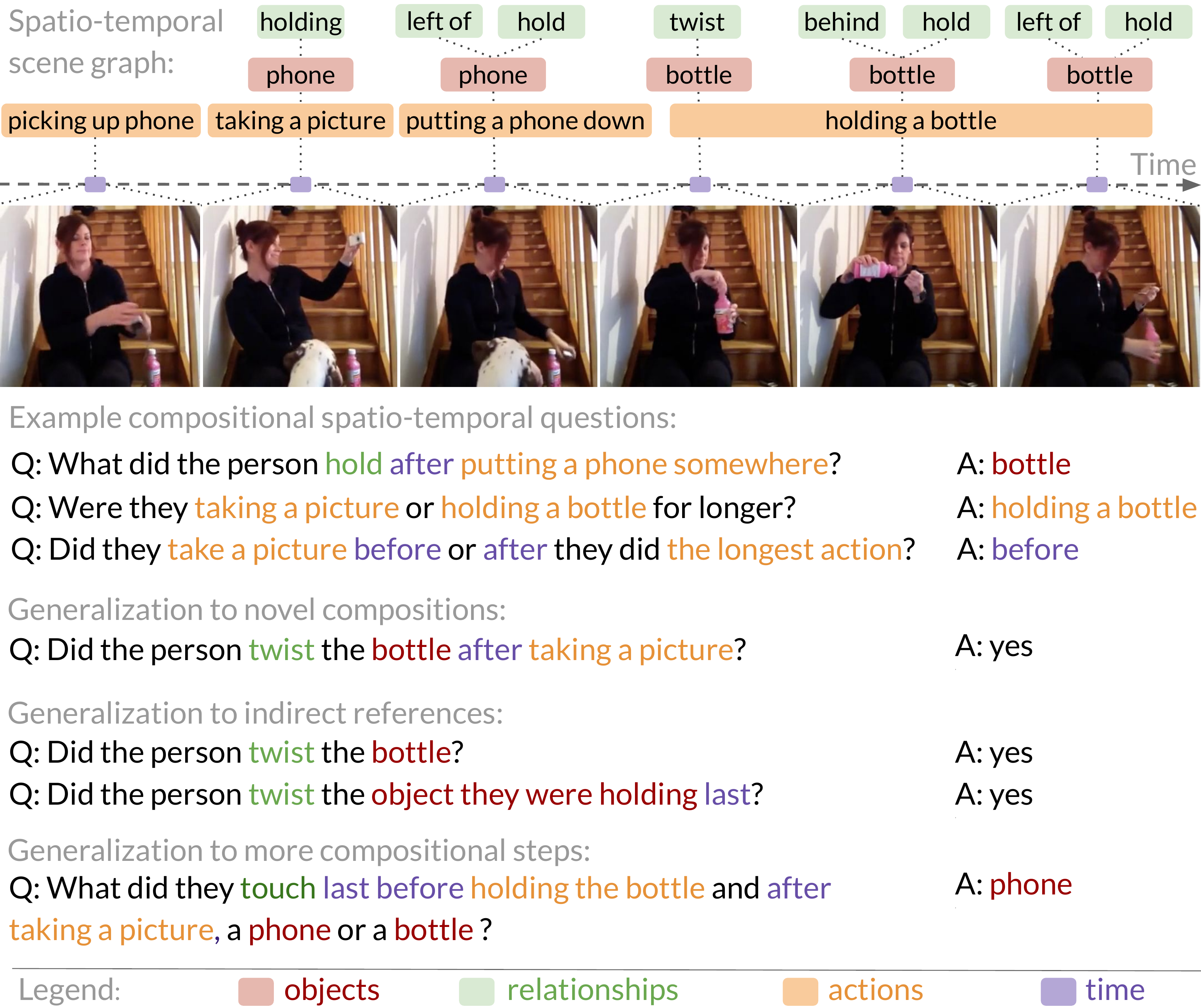}
    \caption{We introduce AGQA: a new benchmark to test for compositional spatio-temporal reasoning. AGQA contains a balanced $3.9M$ and an unbalanced $192M$ question answer pairs associated with $9.6K$ videos. We design handcrafted programs that operate over spatio-temporal scene graphs to generate questions. These questions explicitly test how well models generalize to novel compositions unseen during training, to indirect references of concepts, and to more compositional steps.}
    \label{fig:pull}
\end{figure}

People represent visual events as a composition of temporal actions, where each action encodes how an actor's relationships with surrounding objects evolves over time~\cite{reynolds2007computational,speer2007human,kurby2008segmentation,lillo2014discriminative}. For instance, people can encode the video in Figure~\ref{fig:pull} as a set of actions like \action{putting a phone down} and \action{holding a bottle}. The action \action{holding a bottle} can be further decomposed into how the actor's relationship with the \object{bottle} evolves -- initially the actor may be \relationship{twisting} the \object{bottle} and then later shift to \relationship{holding} it. This ability to decompose events is reflected in the language people use to communicate with one another~\cite{chomsky2002syntactic,montague1970universal},
so tasks involving both vision and language comprehension, such as answering questions about visual input, can test models' compositional reasoning capability.
We can ask questions like ``What did the person \relationship{hold} \temporal{after} \action{putting a phone down}?'' and expect 
a model capable of compositional spatio-temporal reasoning to answer ``\object{bottle}.'' While such behavior seems fundamental to developing vision models that can reason over events, the vision community has only developed compositional question answering benchmarks using static images~\cite{hudson2019gqa} or synthetic worlds~\cite{lake2018generalization,yi2019clevrer} which either are not spatio-temporal or do not reflect the diversity of real-world events.

\input{tables/datasets.tex}

Although questions and visual events are composed of multiple reasoning steps, existing video question answering benchmarks conflate multiple sources of model errors into a single accuracy metric~\cite{jang2017tgif,xu2017video,maharaj2017dataset,zeng2017leveraging,yu2019activitynet}. Consider this stereotypical question-answer pair, Q: `` What does the \object{bear} on the right do \temporal{after} \relationship{sitting}?'' A: ``\relationship{stand up}''~\cite{jang2017tgif}. A model's inability to answer such questions does not afford any deeper insights into the model's capabilities. Did the model fail because it is unable to identify objects like \object{bear} or relationships like \relationship{sitting} or does it fail to reason over the temporal ordering implied by the word \temporal{after}? Or did the model fail for a combination of these reasons?

Not only are failure cases difficult to analyze, but the inputs where the model correctly guesses the answer are equally difficult to dissect. Due to biases in answer distributions and the non-uniform distribution of occurrences of visual events, models may develop ``cheating'' approaches that can superficially guess answers without learning the underlying compositional reasoning process~\cite{li2019repair,yang2020gives}.
To effectively measure how well models jointly compose spatio-temporal reasoning over objects, their relationships, and temporal actions, we need newer benchmarks with more granular control over question composition and the distribution of concepts in questions and answers.

To measure whether models exhibit compositional spatio-temporal reasoning, we introduce Action Genome Question Answering (AGQA)\footnote{Project page: \url{https://tinyurl.com/agqavideo}}. AGQA presents a benchmark of $3.9M$ balanced and $192M$ unbalanced question answer pairs associated with $9.6K$ videos. 
We validate the accuracy of the questions and answers in AGQA using human annotators for at least 50 questions per category and find that annotators agree with $86.02\%$ of our answers.
Each question is generated by a handcrafted program that outlines the necessary reasoning steps required to answer a question. 
The programs that create questions operate over Charades' action annotations and Action Genome's spatio-temporal scene graphs, which ground all objects with bounding boxes and actions with time stamps in the video~\cite{ji2020action,sigurdsson2016hollywood}. These programs also provide us with granular control over which reasoning abilities are required to answer each question. 
For example, some questions in AGQA only require understanding the temporal ordering of actions (e.g.~``Did they \action{take a picture} \temporal{before} or \temporal{after} they did \action{the longest action}?'') while some others require understanding actions in tandem with relationships (e.g.~``What did the person \relationship{hold} \temporal{after} \action{putting a phone somewhere}?'').
We control bias using rejection sampling on skewed answer distributions and across families of different compositional structures.

With our granular control over the question generation process, we also introduce a set of new training/test splits that test for particular forms of compositional spatio-temporal desiderata: generalization to novel compositions, to indirect references, and to more compositional steps. We test whether models (PSAC, HME, and HRCN~\cite{fan2019heterogeneous,le2020hierarchical,li2019beyond}) generalize to novel compositions unseen during training --- the training set can contain the relationship \relationship{twist} and the object \object{bottle} separately while the test set requires reasoning over questions such as ``Did the person \relationship{twist} the \object{bottle} \temporal{after} \action{taking a picture}?'' with both concepts paired together in a novel composition. Similarly, we test whether models generalize to indirect references of objects by replacing objects like \object{bottle} in ``Did the person \relationship{twist} the \object{bottle}?'' with an indirect reference to make the question ``Did the person \relationship{twist} the \object{object they were holding last}?'' Finally,  we test whether models generalize to questions with more reasoning steps by constraining the test set to questions with more reasoning steps than those in the training set (e.g.~``What did they \relationship{touch} \temporal{last before} \action{holding the bottle} but \temporal{after} \action{taking a picture}, a \object{phone} or a \object{bottle}?''). 

Using AGQA, we evaluate modern visual reasoning systems (PSAC, HME, and HRCN~\cite{fan2019heterogeneous,le2020hierarchical,li2019beyond}), and find that they barely perform better than models that purely exploit linguistic bias. The highest performing model achieves only $47.74\%$ accuracy, and HCRN performs only $0.42\%$ better than a linguistic-only version. While there is some evidence that models generalize to indirect references, all of them decrease in accuracy when the number of compositional steps increase and none of them generalize to novel compositions.

%% file: tables/datasets.tex
\begin{table*}[t]
\caption{AGQA is $3$ orders of magnitude larger than all existing VideoQA benchmarks. It contains real-world videos and compositional open-answer questions with action, object, and relationship grounding. AGQA's questions focus on visual comprehension and do not require common sense or dialogue understanding.}
\label{tab:dataset_compare}
\centering
\resizebox{\linewidth}{!}{%
\begin{tabular}{@{\extracolsep{4pt}}l rrc cccr ccc@{}}
 \multirow{2}{*}{Dataset} & \multicolumn{3}{c}{Video} & \multicolumn{4}{c}{Question answers} & \multicolumn{3}{c}{Grounding} \\ 
 \cline{2-4}\cline{5-8}\cline{9-11}
 & Avg. length (s) & \# videos (K) & Real-world & Not dialogue related & Open answer & Compositional & \# questions & objects & relationships & actions \\ 
 \hline
MarioQA~\cite{mun2017marioqa} & 3-6 & 188 &  & \checkmark & \checkmark & & 188K \\
CLEVRER~\cite{yi2019clevrer} & 5 & 20 &  & \checkmark & \checkmark & \checkmark & 282K & \checkmark & \checkmark & \checkmark\\ \hline
Pororo-QA~\cite{kim2017deepstory} & 1.4 & 16.1 & \checkmark &  &  & & 9K \\
MovieQA~\cite{tapaswi2016movieqa}  & 202.7 & 6.77 & \checkmark &  &  & & 6.4K  & & & \checkmark \\
SocialIQ~\cite{zadeh2019social} & 99 & 1.25 & \checkmark &  &  & & 7.5K \\
TVQA~\cite{lei2018tvqa} & 76.2 & 21.8 & \checkmark &  &  & \checkmark & 152.5K &  & & \checkmark \\
TVQA+~\cite{lei2019tvqa} & 7.2 & 4.2 & \checkmark &  &  & \checkmark & 29.4K & \checkmark & & \checkmark \\\hline
MovieFIB\cite{maharaj2017dataset} & 4.9 & 118.5 & \checkmark & \checkmark & \checkmark & & 349K \\ 
TGIF-QA~\cite{jang2017tgif} & 3.1 & 71.7 & \checkmark & \checkmark & \checkmark & & 165.2K \\
MSVD-QA~\cite{xu2017video} & $<$10 & 1.97 & \checkmark & \checkmark & \checkmark & & 50.5K \\
Video-QA~\cite{zeng2017leveraging} & 45 & 18.1 & \checkmark & \checkmark & \checkmark & & 175K \\
MSRVTT-QA~\cite{xu2017video} & 10-30 & 10 & \checkmark & \checkmark & \checkmark & & 243K \\
ActivityNet-QA~\cite{yu2019activitynet} & 180 & 5.8 & \checkmark & \checkmark & \checkmark & & 58K \\
\hline
\textbf{AGQA} & 30 & 9.6 & \checkmark & \checkmark & \checkmark & \checkmark &  \textbf{192M} & \checkmark & \checkmark & \checkmark\\
\end{tabular}}
\end{table*}

%% file: sections/02_related.tex
\section{Related Work}
Our work lies within the field of video understanding using language and is targeted towards the question answering task. We use spatio-temporal scene graphs to generate our questions, and we provide a suite of new evaluation metrics to measure compositional spatio-temporal reasoning. 

\noindent\textbf{Image question answering benchmarks.}
A wide variety of visual question answering benchmarks have been created over the past five years~\cite{johnson2017clevr,hudson2019gqa,antol2015vqa,zellers2019recognition,goyal2017making,krishna2017visual,zhu2016visual7w,kim2020answering}. These benchmarks vary in input, from synthetic datasets~\cite{johnson2017clevr}, to cartoons~\cite{antol2015vqa},  charts~\cite{kim2017deepstory}, or real-world images~\cite{hudson2019gqa,krishna2017visual,zhu2016visual7w,goyal2017making,zellers2019recognition,antol2015vqa}. They also vary in the type of questions asked, from descriptive questions (who, what, where, when, which, why, how)~\cite{zhu2016visual7w}, to ones requiring commonsense reasoning~\cite{zellers2019recognition}, spatial compositional reasoning~\cite{johnson2017clevr,hudson2019gqa}, or spatial localization~\cite{zhu2016visual7w,krishna2017visual,hudson2019gqa}. These benchmarks facilitated the development of many model architectures and learning algorithms that demonstrate spatial compositional reasoning abilities~\cite{lu2016hierarchical,vatashsky2020vqa,chen2020counterfactual}. However, none of these measure temporal reasoning beyond guessing common sense actions that usually require external knowledge~\cite{zellers2019recognition}. 

\noindent\textbf{Video question answering benchmarks.}
As shown by the benchmarks in Table~\ref{tab:dataset_compare}, there is a growing interest in measuring video reasoning capabilities using question answering~\cite{tapaswi2016movieqa,lei2018tvqa,jang2017tgif,kim2017deepstory,xu2017video,maharaj2017dataset,zeng2017leveraging,yu2019activitynet,yi2019clevrer}. Several of these prominent benchmarks rely on dialogue and plot summaries instead of a video's visual contents~\cite{lei2018tvqa,tapaswi2016movieqa,kim2017deepstory,zadeh2019social}, resulting in models with a stronger dependence on the dialogue than on the visual input and therefore reducing the benchmark's effectiveness at measuring visual reasoning~\cite{tapaswi2016movieqa,lei2018tvqa}. 

\begin{figure}[t]
    \centering
    \includegraphics[width=\columnwidth]{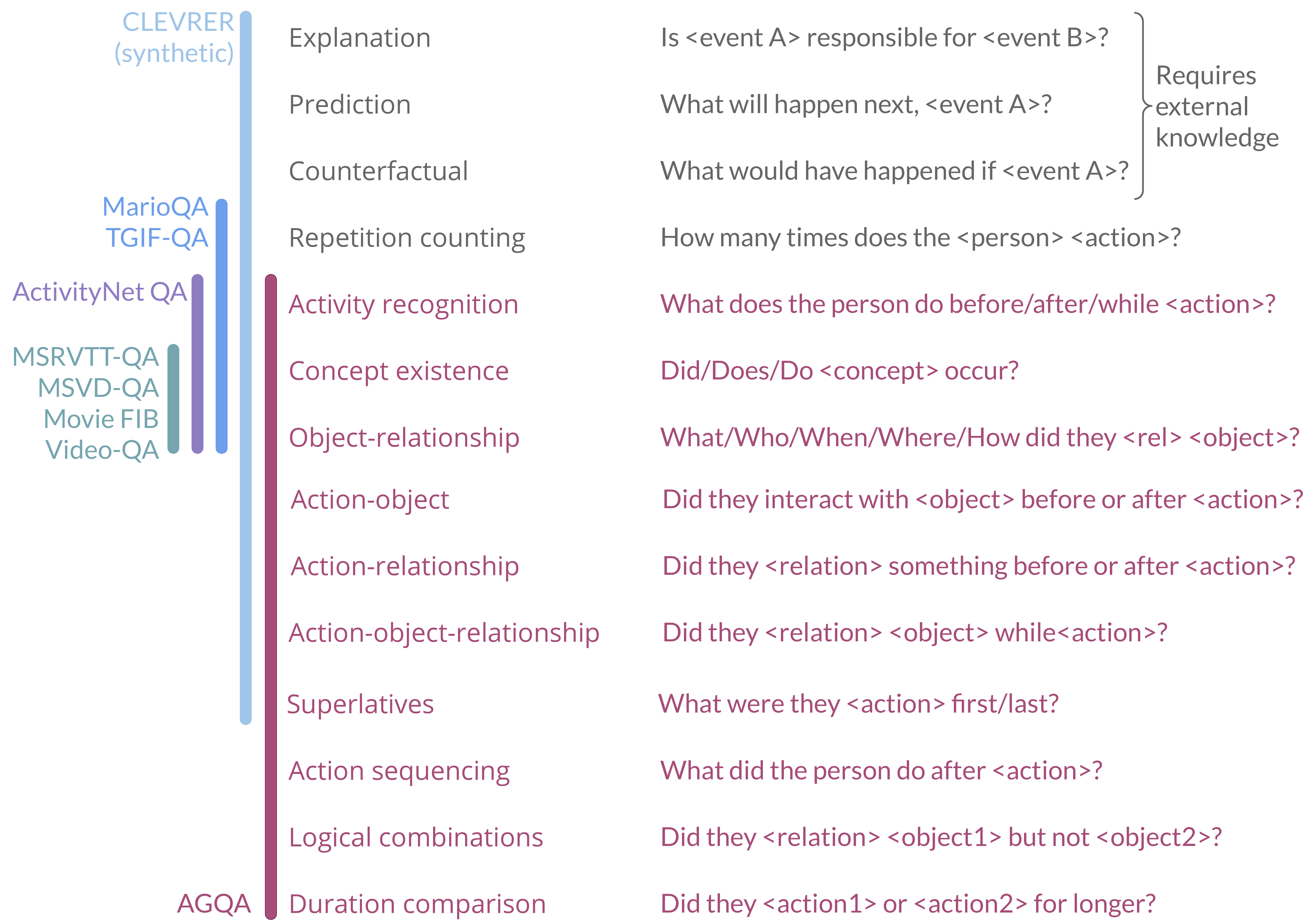}
    \caption{AGQA contains a variety of compositional spatio-temporal reasoning types that are absent from existing real-world video-only corpuses, including duration of actions, interactions between relationships and actions, action sequencing, and logical combinations. We focus on questions that require visual understanding, so we do not have questions that require external knowledge.}
    \label{fig:questions}
\end{figure}

Some video-only question-answering benchmarks are synthetically generated~\cite{yi2019clevrer, mun2017marioqa}, which affords the granular control necessary to measure model abilities like causality~\cite{yi2019clevrer}, or counting~\cite{mun2017marioqa}. However, these benchmarks use short video clips, utilize only a handful of objects, focus on questions that require commonsense or external knowledge (Figure~\ref{fig:questions}), and lack the visual diversity of real-world videos. Other video-only benchmarks suffer from the biases and simplicity associated with human generated questions~\cite{yu2019activitynet,tapaswi2016movieqa,jang2017tgif,lei2018tvqa} or descriptions~\cite{xu2017video,zeng2017leveraging}. The largest human-annotated~\cite{lei2018tvqa} and generated~\cite{maharaj2017dataset} datasets contain $152.5K$ and $349K$ questions. In comparison, our corpus is purely vision based, is three orders of magnitude larger, and evaluates complex and multi-step reasoning.

\noindent\textbf{Scene graphs.}
Scene graphs were first introduced as a Cognitive Science~\cite{biederman1982scene,wolfe1998visual} inspired representation for static images~\cite{krishna2017visual,johnson2015image}. Each scene graph encodes objects as nodes in the image and pairwise relationships between objects as directed edges connecting nodes. 
The Computer Vision community has utilized the scene graph representation for a variety of tasks including visual question answering~\cite{johnson2017inferring}, relationship modeling~\cite{lu2016visual}, object localization~\cite{krishna2018referring}, evaluation~\cite{anderson2016spice}, generation~\cite{johnson2018image,ashual2019specifying}, retrieval~\cite{ashual2019specifying,johnson2015image} and few-shot learning~\cite{chen2019scene,dornadula2019visual}. Of particular interest to our project is how scene graphs from Visual Genome~\cite{krishna2017visual} were used to create GQA, a benchmark for compositional spatial reasoning over an image~\cite{hudson2019gqa}. Our work is a generalization of GQA's pipeline. While GQA uses indirect references to objects with attributes (e.g. ``red'') and spatial relations (e.g. \relationship{to the left of}), we also use temporal localizations (e.g.~\temporal{before}), indirect action references (e.g.~\action{the longest action}), and changes in a subject's relationship with objects over time (e.g.~\temporal{before} \relationship{holding} the \object{dish}). Our programs operate over Action Genome's spatio-temporal scene graphs to automatically generate question-answer-video pairs~\cite{ji2020action}.

\begin{figure*}[t]
    \centering
    \includegraphics[width=\linewidth]{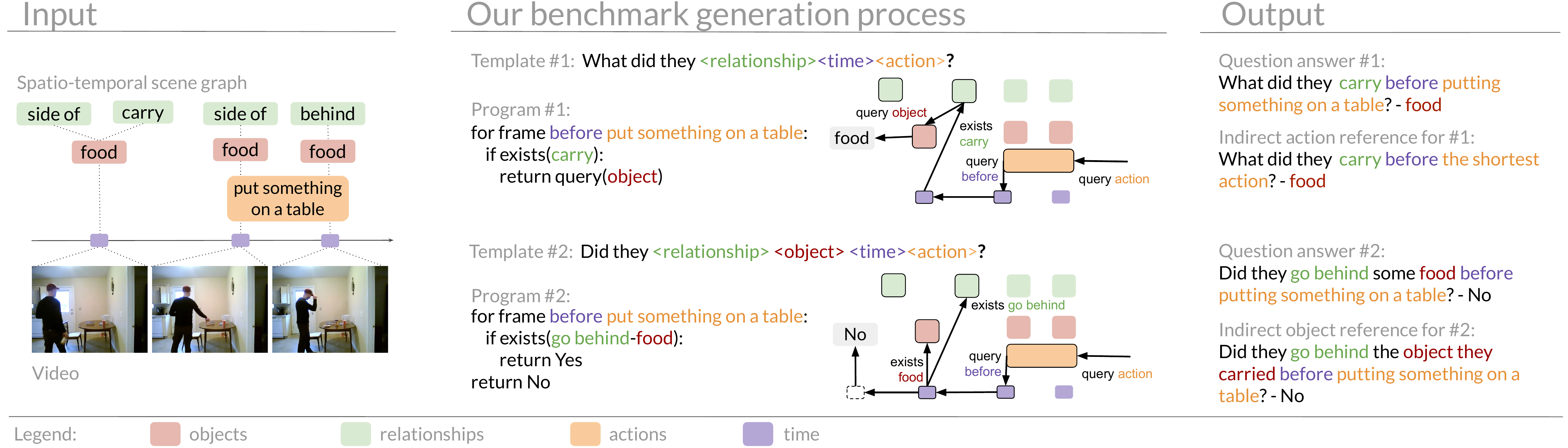}
    \caption{(Left:) Our benchmark generation process expects a dataset of videos with spatio-temporal scene graphs as input. (Middle:) We handcraft programs that operate over the scene graphs to generate questions and answers. (Right:) We balance generated questions and their corresponding answers using rejection sampling to avoid biases that models can exploit. We can control the number of reasoning steps required to answer a question by either developing more complex programs or by referencing visual concepts using indirect references (e.g. referring to a specific action as \action{the shortest action} or object as \object{the object they carried}).
    }
    \label{fig:system}
\end{figure*}

\noindent\textbf{Compositional reasoning.}
While there are numerous definitions of compositionality, we in particular use what is more colloquially referred to as bottom-up compositionality --- ``the meaning of the whole is a function of the meanings of its parts''~\cite{cresswell1973logics}. In our case, reasoning about the question ``Was the person \action{running} or \action{sitting} for longer?'' requires finding the start and end of when the person was \action{running} and \action{sitting}, subtracting the start from the end, then comparing the resulting lengths.
Unfortunately, the most popular benchmarks and metrics defined to study compositional behavior have been limited to synthetic environments~\cite{keysers2019measuring,lake2018generalization,johnson2017clevr,yi2019clevrer} or to static images~\cite{hudson2019gqa}. Recent work has argued the importance of compositionality in enabling models to generalize to new domains, categories, and logical rules~\cite{lake2018generalization,vatashsky2020vqa} and has discovered that current models struggle with multi-step reasoning~\cite{fan2019heterogeneous}. These studies motivate a benchmark like ours that defines multiple metrics to explore compositional reasoning in real-world videos.

%% file: sections/03_method.tex
\section{The AGQA benchmark}

Our benchmark generation process takes videos with annotated spatio-temporal scene graphs~\cite{ji2020action} as input and produces a balanced corpus of question-answer pairs (Figure~\ref{fig:system}). First, we consolidate and augment Action Genome's spatio-temporal scene graphs~\cite{ji2020action} and Charades' action localizations~\cite{sigurdsson2016hollywood} into a symbolic video representation. Next, we handcraft programs that operate over the augmented spatio-temporal scene graphs and generate questions using probabilistic grammar rules.
Then, we reduce biases in answer distributions and by question structure types, resulting in a balanced benchmark that is more robust against ``cheating.'' 
Finally, we create new evaluation metrics that allow us to test how well models generalize to novel compositions, to indirect references, and to more compositional steps.

\begin{figure*}[t]
    \centering
    \includegraphics[width=0.95\linewidth]{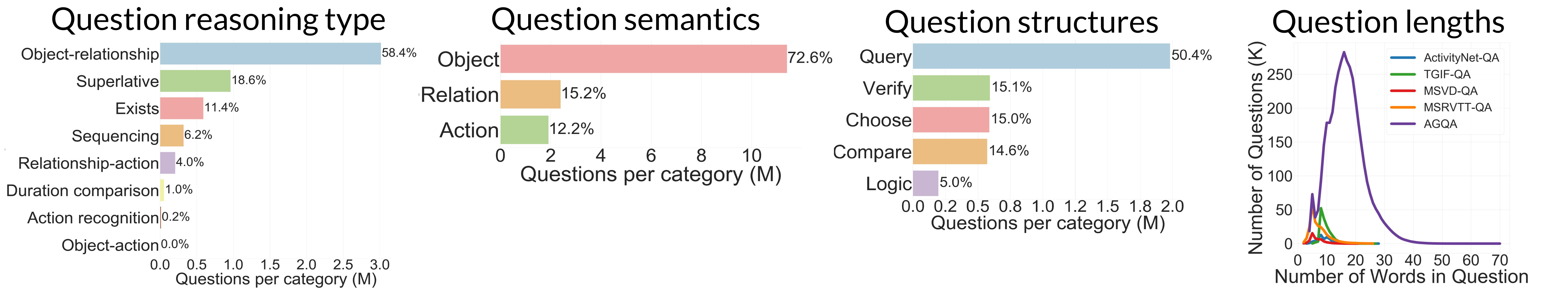}
    \caption{We classify each question in AGQA in three category types. Reasoning types distinguish which reasoning steps are required to answer the question. Semantic types split questions based on whether they are asking about an object, relationship, or action. Question structure types indicate the question's form. We also compare the distribution of question lengths of AGQA and existing video question answering benchmarks.}
    \label{fig:stats}
\end{figure*}

\subsection{Augmenting spatio-temporal scene graphs}
AGQA is generated using programs that operate over Action Genome's spatio-temporal scene graphs. Each spatio-temporal scene graph is associated with a video and contains objects (e.g.~\object{food}, \object{bottle}) that are grounded in video frames, and the spatial relationships (e.g.~\relationship{above}, \relationship{behind}), and contact relationships (e.g.~\relationship{carry}, \relationship{wipe}) that describe an actor's interactions with the objects~\cite{ji2020action}. We augment Action Genome's spatio-temporal scene graphs with actions (e.g.~\action{running}) from the Charades dataset, localized using time stamps for when the action starts and ends~\cite{sigurdsson2016hollywood}.

To use these scene graphs for question generation, we augment them by specifying entailments between actions and relationships, incorporating prior knowledge about action sequencing, merging synonymous annotations, and removing attention relationships. 
Some actions and relationships, such as \action{carrying a blanket} and \relationship{twisting} the \object{blanket}, entail other relationships such as \relationship{holding} and \relationship{touching}. We augment the scene graphs with such entailment relationships to avoid generation of degenerate questions like ``Were they \relationship{touching} the \object{blanket} \temporal{while} \action{carrying the blanket}?'' 
We created heuristics that adjust the start and end times of actions to avoid logical errors. For example, the action \action{taking a pillow from somewhere} would often end after the next action, \action{holding a pillow}, would start. To be able to generate questions that reason over the temporal ordering of these events, we modified the events so that the first action ends before the next one starts. To avoid generating simple questions with only one answer, we use co-occurrence statistics to prune relationships that only occur with one object category (e.g.~\relationship{turning off} a \object{light}). We also consolidate references to similar objects and actions (e.g.~\action{eating a sandwich} and \action{eating some food}) so that each concept is represented by one phrase.
Finally, we remove all attention relationships (e.g.~\relationship{looking at}) from Action Genome's annotations because our human evaluations indicated that evaluators were unable to accurately discern the actor's gaze.

The resulting spatio-temporal scene graphs have more clean, unified, and unambiguous semantics. Our final ontology uses $36$ objects, $44$ relationships, and $157$ actions. There are $7,787$ training and $1,814$ test set scene graphs.

\subsection{Question templates}

To generate question and answer pairs from spatio-temporal scene graphs, we handcraft a suite of programs, each associated with a template (see Figure~\ref{fig:system}). Each template has a variety of natural language question frames that can be filled in by scene graph content. For example, a template ``What did they $<$\relationship{relationship}$><$\temporal{time}$><$\action{action}$>$?'' can generate questions like ``What did they \relationship{tidy} \temporal{after} \action{snuggling with a blanket}?'' and ``What did they \relationship{carry} \temporal{before} \action{putting something on a table}?'' To answer this question, the associated program finds the action \action{put something on a table}, attends to events \temporal{before} that action, finds where the relationship \relationship{carry} occurs, and finally queries for the object.

This generation process associates each question with the reasoning skills and number of reasoning steps used to answer it. While some of the spatio-temporal reasoning skills required to answer our questions are inspired from existing corpora, successfully answering AGQA's questions requires a variety of new spatio-temporal reasoning absent in existing benchmarks (see Figure~\ref{fig:questions}). Along with incorporating more reasoning skills, we increase the number of compositional reasoning steps necessary to answer a question by allowing question templates to use phrases that localize a time within the video and indirect references to objects, relationships, and actions. For example, we can replace \object{food} with the indirect reference \object{the object being carried} or \action{walking through a doorway} with \action{the shortest action}.

For each question, we also keep track of its answer type, semantic class, and structure. Open answer questions have many possible answers, while binary questions have answers that are Yes/No, \temporal{before}/\temporal{after}, or are specified as one of two options (e.g.~\relationship{carrying} or \relationship{throwing}) within the question.  A question's semantic class describes its main subject, a (1) object; (2) relationship; or (3) action. AGQA classifies questions into five structure categories: (1)  query for all open questions; (2) compare for comparisons (3) choose for questions that present two alternatives from which to choose; (4) verify questions that respond yes or no to the question's contents; and (5) logic questions with logical conjunctions. We display the distribution of questions across these categories in Figure~\ref{fig:stats}.

Before adding a question to the benchmark, we ensure that there is no ambiguity in answers by removing questions for which multiple elements could satisfy the constraints of the question.
We avoid nonsensical compositions (e.g.~``Were they \relationship{eating} a \object{mirror}?'') by only asking about object-relationship pairs that occur at least $10$ times in Action Genome. We also delete questions that answer themselves (e.g.~``What did they \relationship{hold} while \action{holding a blanket}?''). Finally, we remove questions that always have one answer across all our videos (e.g.~``Are they \relationship{wearing} \object{clothes}?'').

We handcraft $269$ natural language question frames that can be answered from a set of $28$ programs. Using these programs, we generate $192M$ question-answer pairs, with over $45M$ unique questions and $174$ unique answers.

\input{tables/overall_results}

\subsection{Balancing to minimize bias} \label{balancing}

Machine learning models are notoriously adept at exploiting imbalances in question answering datasets~\cite{goyal2017making,hudson2019gqa,johnson2017clevr}. 
We mitigate inflated accuracy scores by balancing our benchmark's answer distributions for each reasoning category and by the distribution of question structures.

We balance answer distributions with an approach inspired by the method described in GQA~\cite{hudson2019gqa}. We first balance all answer distributions for each overall reasoning type and then for each concept within that reasoning type. For example, we first balance the answer distribution for the ``exists'' category, then that of the ``exists-\relationship{taking}-\object{dish}-and-\object{picture}'' category. For binary questions, we ensure that each answer is equally likely to occur. For open answer questions, we iterate over the answers in decreasing frequency order, and re-weight the head of the distribution up to the current iteration to make it more comparable to the tail.

Second, we use rejection sampling to normalize the distribution of question structures. 
Our templates generate more binary questions than the more difficult query questions. We balance the benchmark such that query questions constitute at least $50\%$ of the benchmark. We further balance the binary answer questions such that approximately $15\%$ are comparisons, $15\%$ are choose questions, $15\%$ are verify questions, and $5\%$ use a logical operator. This new distribution of question structures increases the benchmark's difficulty and makes the distribution of required reasoning skills more varied.

Our balancing procedure reduced AGQA from an unbalanced set of $192M$ question answer pairs to a balanced benchmark with $3.9M$ question answer pairs. We provide a detailed algorithm in supplementary materials.

\subsection{New compositional spatio-temporal splits}
With control over our generated set of questions, we measure how well models perform across different reasoning skills, semantic classes, and question structures. We also introduce a new set of train/test splits to test for particular forms of compositional spatio-temporal reasoning that require generalization to novel and more complex concepts.

\noindent\textbf{Novel compositions:} To test whether models can disentangle distinct concepts and combine them in novel ways, we hand-select a set of concept pairs to only appear in the test set. For example, we remove all training questions that contain the phrase \temporal{before} \action{standing up}, but retain only questions with the specified phrases in the test set.

\noindent\textbf{Indirect references:} The semantic categories in a question can be referred to directly (e.g.~\object{blanket}, \relationship{holding}, and \action{eating something}) or indirectly (e.g.~\object{the object they threw}, \relationship{the thing they did to the laptop}, and \action{the longest action}). Indirect references make up the core method through which we increase compositional steps. This metric compares how well models answer a question with indirect references if they can answer it with the direct reference.

\noindent\textbf{More compositional steps:} To test whether models generalize to more compositional steps, we filter the training set to contain simpler questions with $\leq M$ compositional steps, such as ``What did they \relationship{touch}?'' then reduce the test set to contain only questions with $>M$ compositional steps, such as ``What did they \relationship{touch} \temporal{last before} \action{holding the bottle} but \temporal{after} \action{taking a picture}, a \object{phone} or a \object{bottle}?''

%% file: tables/overall_results.tex
\begin{table*}
\caption{Although humans verify $86.02\%$ of our answers as correct, modern vision models struggle on a variety of different reasoning skills, semantic classes, and question structures. In fact, most of the increase in HCRN's performance comes from exploiting linguistic biases instead of from visual comprehension.}
\label{tab:global}
\centering
\resizebox{\linewidth}{!}{%
\begin{tabular}{rrrrrrrr}
                           & Question Types       & Most Likely & PSAC~\cite{li2019beyond} & HME~\cite{fan2019heterogeneous} & HCRN (w/o vision)\cite{le2020hierarchical} & HCRN\cite{le2020hierarchical} & Human \\ \hline
\multirow{8}{*}{\rotatebox{90}{Reasoning}} & object-relationship  & 8.82        & 34.75                                          & \textbf{43.91}                                                 & 42.33                                                             & 43.00                                                & 80.65 \\
                           & relationship-action  & 50.00       & 56.84                                          & 57.84                                                 & \textbf{58.06}                                                            & 56.75                                                & 90.20 \\
                           & object-action        & 50.00       & 58.33                                          & 50.00                                                 & 51.67                                                             & \textbf{63.33}                                                & 93.75 \\
                           & superlative          & 10.29       & 30.51                                          &\textbf{41.10}                                                 & 36.83                                                             & 37.48                                                & 81.25 \\
                           & sequencing           & 49.15       & 59.95                                          & 59.60                                                 & \textbf{62.11}                                                             & 61.28                                                & 90.77 \\
                           & exists               & 50.00       & 69.94                                          & 70.01                                                 & 72.12                                                             & \textbf{72.22}                                                & 79.80 \\
                           & duration comparison  & 23.70       & 29.75                                          & 44.19                                                 & \textbf{45.24}                                                             & 45.10                                                & 92.00 \\
                           & activity recognition & 4.72        & 3.78                                           & 3.23                                                  & 7.57                                                              & \textbf{11.21}                                                & 78.00 \\ \hline
\multirow{3}{*}{\rotatebox{90}{Semantic}}  & object               & 9.38        & 32.79                                          & \textbf{42.48}                                                 & 40.74                                                             & 41.55                                                & 87.97 \\
                           & relationship         & 50.00       & 65.51                                          & 66.10                                                 & \textbf{67.40}                                                             & 66.71                                                & 83.58 \\
                           & action               & 32.91       & 57.91                                          & 58.12                                                 & \textbf{60.95}                                                             & 60.41                                                & 86.45 \\ \hline
\multirow{5}{*}{\rotatebox{90}{Structure}} & query                & 11.76       & 27.20                                          & 36.23                                                 & 36.50                                                             & \textbf{37.18}                                                & 83.53 \\
                           & compare              & 50.00       & 56.68                                          & 58.06                                                 & \textbf{59.65}                                                             & 58.77                                                & 92.53 \\
                           & choose               & 50.00       & 33.41                                          & \textbf{49.32}                                                 & 39.52                                                             & 40.60                                                & 83.02 \\
                           & logic                & 50.00       & 67.48                                          & 69.75                                                 & 69.47                                                             & \textbf{69.90}                                                & 70.69 \\
                           & verify               & 50.00       & 68.34                                          & 68.40                                                 & 70.94                                                             & \textbf{71.09}                                                & 88.26 \\ \hline
\multirow{3}{*}{\rotatebox{90}{Overall}}   & binary               & 50.00       & 54.19                                          & \textbf{59.77}                                                 & 57.93                                                             & 58.11                                                & 86.65 \\
                           & open                 & 11.76       & 27.20                                          & 36.23                                                 & 36.50                                                             & \textbf{37.18}                                                & 83.53 \\
                           & all                  & 10.35       & 40.40                                          & \textbf{47.74}                                                 & 47.00                                                             & 47.42                                                & 86.02
\end{tabular}
}
\end{table*}

%% file: sections/04_experiments.tex
\section{Experiments and analysis}
We begin our experiments with scores from a human validation task on the AGQA benchmark that evaluates the correctness of our benchmark generation process. Next, we compare state-of-the-art question answering models on AGQA, revealing a large gap between model performance and human validation of our dataset. We report how well models perform on spatio-temporal reasoning, for different semantics, and for each structural category. Finally, we report how well models generalize to novel compositions, to indirect references, and to more compositional steps. All experiments run on the balanced version of AGQA.

\noindent\textbf{Models: } 
We evaluate three recent video question answering models: PSAC~\cite{li2019beyond}, HME~\cite{fan2019heterogeneous}, and HCRN~\cite{le2020hierarchical}. PSAC uses positional self-attention and co-attention blocks to integrate visual and language features~\cite{li2019beyond}. HME builds memory modules for visual and question features and then fuses them together~\cite{fan2019heterogeneous}. HCRN, a current best model, stacks a reusable module into a multi-layer hierarchy, integrating motion, question, and visual features at each layer~\cite{le2020hierarchical}. We use identical feature representations, from the ResNet \textit{pool5} layer and ResNeXt-101, for all models.

We compare performance against a ``Most-Likely'' baseline that reports the accuracy of always guessing the most common answer after balancing (Section~\ref{balancing}). 
Binary questions have a Most-Likely accuracy of $50\%$ because they ask a Yes/No or before/after question, or they list answers in the question (e.g.~``What did they \relationship{hold}, a \object{bag} or a \object{dish}?'').

\subsection{Human evaluation}
To quantify the errors induced by our benchmark generation process, we hire subjects at a rate of $\$15$/hr in accordance with fair work standards on Amazon Mechanical Turk~\cite{whiting2019fair}. We present at least $50$ randomly sampled question per question type from AGQA to our subjects. We used the majority vote of three subjects as the final human answer.
Human validation labeled $86.02\%$ of our answers as correct, implying that about $13.98\%$ of our questions contain errors. These errors originate in scene graph annotation errors and ambiguous relationships.  We describe in supplementary materials the sources of human error and a second validation task.
To put this number in context, GQA~\cite{hudson2019gqa} and CLEVR~\cite{johnson2017clevr}, two recent automated benchmarks, report $89.30\%$ and $92.60\%$ human accuracy, respectively.

\input{tables/compositionality_results}

\input{tables/novel_comp_results}

\subsection{Performance across reasoning abilities}\label{sec:reasoning}

Each question is associated with the one or more reasoning abilities necessary to answer the question. By analyzing performance on each reasoning category, we get a detailed understanding of each model's reasoning skills. Overall, we find that across the different reasoning categories HME and HCRN perform better than PSAC (Table~\ref{tab:global}). HME outperforms the others on questions asking about superlatives, while HCRN outperforms the others on questions involving sequencing and activity recognition. 

However, for most reasoning categories, HCRN does not outperform a language-only version of itself (HCRN w/o vision) by more than $1.5\%$. In fact, HCRN performs worse than its language-only counterpart on questions that involve sequencing actions as well as questions that reason over the length of time actions occurred. The only two reasoning categories in which the HCRN model outperforms the language-only baseline by more than $1.5\%$ are on questions that focus on activity recognition and on questions comparing object-action interaction. Although HCRN improves on questions that require activity recognition, these questions are very challenging for all models and for humans. A more detailed breakdown of each section split by binary and open question types is in supplementary materials.

\subsection{Performance across question semantics}

We also compare how models perform across different question semantic categories (Table~\ref{tab:global}).
HCRN only improves over the language-only variant for questions that revolve around objects. However, object-related questions were the most difficult for all three models. 

\subsection{Performance across question structures}

Different question structures also appear more challenging than others (Table~\ref{tab:global}). Open-ended query questions are very challenging and have the lowest accuracy for all models. HCRN outperforms the language-only variant in this category by only $0.68\%$. The models have similar performances for each structural category, with the exceptions that PSAC struggles the most with open-ended questions and HME outperforms the rest on choosing questions.

\input{tables/indirects}

\subsection{Generalization to novel compositions}
All models struggle when tested on novel compositions unseen during training (Table~\ref{tab:compo}). HME outperforms the others overall and on binary questions, while HCRN performs best for open-ended questions. However, no model performs much better than the Most-Likely model on open questions. Only HME outperforms $50\%$ on binary questions with $52.39\%$ accuracy.

We further break down the performance on novel compositions by composition type (Table~\ref{tab:novel}). For example, in the sequencing category we remove compositions like \temporal{before} \action{standing up} from the training set and test how well models perform on questions with those compositions in the test set. 
We find that models perform the worst on novel compositions that involve new object and relationship pairs and best on reasoning about the length of novel actions. HME generalizes best to novel sequencing and superlative compositions, while HCRN generalizes best to novel compositions of the duration of actions and object-relationship pairs. 

 \begin{figure}[t]
     \centering
     \includegraphics[width=1\linewidth]{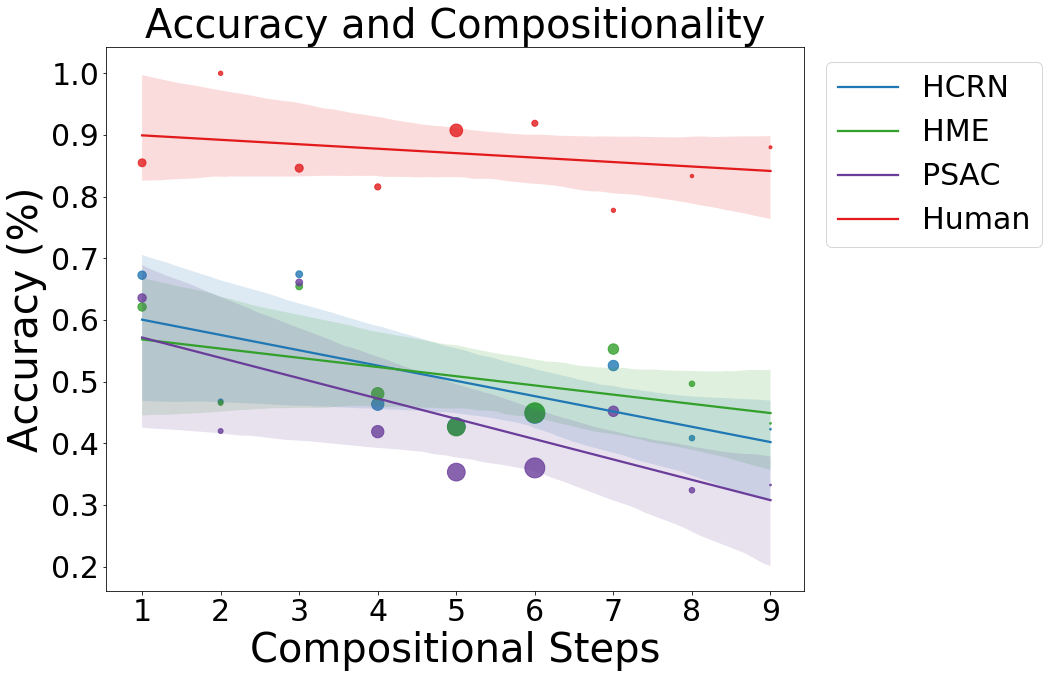}
     \caption{For all three models, we fit a linear regression and find that accuracy is negatively correlated with the number of compositional reasoning steps used to answer the question. However, the $R^2$ scores are relatively weak for all three: HCRN ($.43$), HME ($.24$), and PSAC ($.51$). This is likely because all three models barely outperform the Most-Likely baseline, even for small compositional steps. The human validation study's $R^2$ score is $.09$. The size of the dots correlates with the number of questions with that many steps, with the model's test set size scaled to be $1000$x smaller. The shaded area is the $80\%$ confidence interval.
     }
     \label{fig:compo_fig}
 \end{figure}

\subsection{Generalization to indirect references}

We report precision and recall for how well models generalize to indirect references in Table~\ref{tab:indirect}. 
HCRN generalizes best to indirect object and temporal references, while HME generalizes best to relationship and action indirect references. However, the models still fail on at least nearly a fifth of questions with indirect references, even when it correctly answers the direct counterpart.

\subsection{Generalization to more compositional steps}\label{sec:steps_exp}

When trained on simple questions and tested on questions with more compositional steps, the models outperform the Most-Likely baseline on open questions. However, they still achieve less than $50\%$ accuracy on binary questions. 
HCRN performs the best on open-ended questions, but HME generalizes better overall to questions with more compositional steps than the other models.
This is likely because HME's architecture was explicitly designed to answer semantically complex questions, as it has a memory network for reasoning over question features~\cite{fan2019heterogeneous}. 

Despite some aptitude at generalizing to more complex questions, these models' accuracy scores decrease as the number of compositional steps increase (Figure~\ref{fig:compo_fig}).

%% file: tables/compositionality_results.tex
\begin{table}[t]
\centering
\caption{We introduce new training/test splits to measure whether models generalize to novel compositions and to more compositional steps. B and O refer to binary and open questions. Overall, none of the models generalize.}
\label{tab:compo}
\resizebox{\linewidth}{!}{
\begin{tabular}{rrrrrr}
\multicolumn{1}{l}{}                      & \multicolumn{1}{l}{} & \multicolumn{1}{c}{Most Likely} & \multicolumn{1}{c}{PSAC} & \multicolumn{1}{c}{HME} & \multicolumn{1}{c}{HCRN} \\ \hline
Novel        & B                    & 50.00                           & 43.00                    & \textbf{52.39}          & 43.40                    \\
 Composition                                         & O                    & 15.87                           & 14.80                    & 19.46                   & \textbf{23.72}           \\
                                          & All                  & 10.55                           & 32.49                    & \textbf{40.11}          & 36.06                    \\ \hline
More & B                    & 50.00                           & 35.39                    & \textbf{48.09}          & 42.46                    \\
 Compositional                                         & O                    & 14.51                           & 28.00           & 33.47                   & \textbf{34.81}                    \\
 Steps                                         & All                  & 12.81                           & 31.13                    & \textbf{39.70}          & 38.00                   
\end{tabular}
}
\end{table}

%% file: tables/novel_comp_results.tex
\begin{table}[t]
\caption{Here we break down the models' accuracy when generalizing to novel compositions of different reasoning types.}
\label{tab:novel}
\centering
\resizebox{\linewidth}{!}{
\begin{tabular}{rrrrr}
\multicolumn{1}{l}{} & \multicolumn{1}{l}{Most Likely} & \multicolumn{1}{l}{PSAC} & HME            & \multicolumn{1}{l}{HCRN} \\ \hline
Sequencing           & 13.67                           & 38.35                    & \textbf{44.77} & 42.91                    \\
Superlative          & 12.60                           & 31.97                    & \textbf{41.48} & 34.01                    \\
Duration             & 10.96                           & 38.65                    & 48.19          & \textbf{48.90}           \\
Obj-rel              & 35.63                           & 19.12                    & 22.17          & \textbf{25.71}          
\end{tabular}
}
\end{table}

%% file: tables/indirects.tex
\begin{table}[t]
\centering
\caption{We evaluate performance on questions with indirect references. Precision values are accuracy on these indirect questions when the corresponding question with only direct references was answered correctly, while recall values are accuracy on all questions with that kind of indirect reference.}
\label{tab:indirect}
\resizebox{\linewidth}{!}{
\begin{tabular}{rrrrrrr}
\multicolumn{1}{l}{} & \multicolumn{2}{c}{PSAC} & \multicolumn{2}{c}{HME}         & \multicolumn{2}{c}{HCRN}        \\
\multicolumn{1}{c}{} & Precision    & Recall    & Precision      & Recall         & Precision      & Recall         \\ \hline
Object               & 64.82        & 38.64     & 79.16          & \textbf{47.32} & \textbf{81.03} & 46.29          \\
Relationship         & 40.84        & 24.12     & \textbf{48.6}  & 29.39          & 46.77          & \textbf{29.82} \\
Action               & 64.53        & 34.62     & \textbf{81.68} & \textbf{45.15} & 80.22          & 43.05          \\
Temporal             & 66.48        & 33.15     & 80.71          & \textbf{42.91} & \textbf{83.92} & 42.13     
\end{tabular}
}
\end{table}

%% file: sections/05_discussions.tex
\section{Discussion and future work}

In conclusion, we contribute AGQA, a new real-world compositional spatio-temporal benchmark that is $3$ orders of magnitude larger than existing work. Compositional reasoning is fundamental to understanding visual events~\cite{kurby2008segmentation,lillo2014discriminative} and has been sought after recently by a number of papers~\cite{stone2017teaching,ye2019compositional,yu2019compositional,gan2017semantic,kato2018compositional}. However, to the best of our knowledge, AGQA is the first benchmark to use language to evaluate visual compositional desiderata: generalization to novel compositions, to indirect references, and to more compositional steps. Our experiments paint a grim picture --- modern visual systems barely perform better than variants that exploit linguistic bias, and no models generalize to novel compositions. Although these models demonstrated some capability to generalize to more compositional steps, the overall trend was negative; model accuracy decreased as the number of reasoning steps increased.

While the results may appear grim, they also suggests multiple directions for future work to pursue. We expect researchers to utilize AGQA as a benchmark to make progress in the following directions:

\noindent\textbf{Neuro-symbolic and semantic parsing approaches:} We believe that the fundamental component missing in current models is the ability to extract systematic rules from the training questions. A model might perform better if it can operate in the ``rule-space'' using an explicit representation, either using neuro-symbolic~\cite{mao2019neuro} or semantic parsing~\cite{johnson2017inferring,hu2017learning} to convert a question into an executable program.
As AGQA provides ground truth scene graph annotations for all questions, it naturally leads into this line of work. 

\noindent\textbf{Meta-learning and multi-task learning:} Since none of the models exhibited generalization to novel compositions, meta-learning might be a promising objective, which requires models to discover shared underlying compositional rules~\cite{finn2017model}. Such an approach can expose models to a number of learning environments with varying sets of novel compositions during the meta-train step. 
Another formulation worth exploring is multi-task learning, where models also simultaneously learn to detect objects, classify relationships, and recognize actions~\cite{caruana1997multitask}.

\noindent\textbf{Memory and attention based approaches:} HME outperformed other models in generalizing to more compositional steps. Perhaps this improvement is due to its explicit usage of memory when processing the question features. Future work can explore methods to keep track of each reasoning step with memory networks ~\cite{weston2014memory}, or even use attention based approaches~\cite{hudson2018compositional} to iteratively reason over the steps outlines in a question.

AGQA contributes a benchmark evaluating compositional spatio-temporal reasoning in visual systems along a variety of dimensions. The structure of this benchmark provides the computer vision community with multiple directions for future work. 

\textbf{Acknowledgements.} This work was partially supported by the CRA DREU program, the Stanford HAI Institute, and the Brown Institute. We also thank Michael Bernstein, Li Fei-Fei, Lyne Tchapmi, Edwin Pan, and Mustafa Omer Gul for their valuable insights.

%% file: sections/06_supplementary.tex
\setcounter{section}{5}
\section{Supplementary}

The following sections cover more details regarding the results of our experiments and benchmark generation process. First, we separate each model's performance on binary questions that have two possible answers versus on questions that have open answers. Next, we detail how we create high quality questions, including how we augment the spatio-temporal scene graphs, ignore unchallenging and ambiguous questions, and balance answer and question structure distributions. We then explain which compositions were held out when generating the novel composition training/test split. Finally, we describe the two human studies that validate the correctness of our generation process, analyze the source of errors, and provide recommendations for future work.

Although all our experiments were conducted on the balanced version of the dataset, we will also release two additional versions of the benchmark: a full unbalanced version and a smaller subsampled unbalanced version. Since the full unbalanced AGQA is 192M questions, training models with data at this scale might be prohibitive for smaller research groups. The smaller unbalanced set can be used to serve as a benchmark under resource constraints.

\subsection{Binary versus open answer results} \label{bin_open_results}

Since the number and distribution of possible answers for a question affects the model's likelihood of guessing the correct answer, we split the experimental results by binary and open answer questions. Models achieve lower accuracy on open answer questions than they do on binary questions in each reasoning category (Table~\ref{tab:full_global}). 
However, we find that models generalize to indirect references better on open answer questions than on binary questions (Table~\ref{tab:full_indirects}).

We separate our analysis on binary versus open answer questions to appropriately compare each model's performance against the Most-Likely baseline.
The Most-Likely baseline for binary questions (e.g.~Yes/No or \temporal{before}/\temporal{after} answers) is usually higher than that for open answer questions.
In the balanced benchmark, the Most-Likely baseline has a $50\%$ accuracy on Yes/No binary questions. Other binary questions compare the attributes of two elements (e.g. ``Were they \action{fixing a light} or \action{consuming some medicine} for longer?'') or offer a choice between two elements (e.g. ``Did they \relationship{open} a \object{closet} or a \object{refrigerator}?''). All questions with the answer choice have a $50\%$ chance of correctness if the model chooses one of the provided options. However, the category-wide Most-Likely baseline may be lower than $50\%$ because the category-wide Most-Likely answer may not be one of the two presented options in a question. 
For open answer categories, the Most-Likely accuracy baseline is the percent of all questions in the category with the most common answer.

In this section, we explore models' accuracy on binary and open questions in each reasoning, semantic, and structural category (Table~\ref{tab:full_global}). We then further analyze the results from the metrics measuring generalization to novel compositions (Table~\ref{tab:full_novel}),  indirect references (Table~\ref{tab:full_indirects}), and compositional steps (Table~\ref{tab:compo_steps}).

\noindent\textbf{Reasoning categories} (Table~\ref{tab:full_global}): 
Across all reasoning categories, models perform much worse on open answer questions than binary questions. HCRN is generally stronger on open answer questions than the other models, especially for questions involving sequencing, duration, and action recognition. In fact, HCRN improves upon its blind counterpart for all open-ended question categories with the exception of questions involving superlatives, in which it performs only $0.02\%$ worse. In contrast, HCRN performs worse than its blind counterpart for binary questions in the duration, sequencing, and relationship-action categories.

\noindent\textbf{Semantic categories} (Table~\ref{tab:full_global}): 
The non-blind HCRN model outperforms all others on open-ended questions that reason over objects and actions. For binary questions on objects, HME performs over $5\%$ better than all other models, but for questions reasoning over relationships, all models perform within $2\%$ of one another.

\input{tables/full_novel_comp}

\noindent\textbf{Structural categories} (Table~\ref{tab:full_global}):
Each structural class contains either only binary or only open questions, so these results are identical to those in the main paper. For HCRN, using visual features improves accuracy for all structural categories besides compare. 

\noindent\textbf{Novel composition} (Table~\ref{tab:full_novel}): We split element pairs in the novel compositions metric into four different types. We provide more detail on the makeup of each type in Section~\ref{sec:novel_compo}. For each category we pair a phrase with several objects, relationships, or actions to create novel compositions for the test set. For example, the superlative row looks at questions that contain the concept \temporal{first} paired with different relationships, including \relationship{behind} and \relationship{holding}.  

The models struggle to generalize to novel compositions for all categories except duration. They all perform the worst at generalizing to novel object-relationship compositions. 
Across all categories, HCRN performs the best with novel compositions in open-answer questions, but HME performs the best with novel compositions in binary questions.

\noindent\textbf{Indirect references} (Table~\ref{tab:full_indirects}):
This metric measures a model's accuracy on questions with indirect references and phrases that specify one part of the video. The Recall score shows the overall accuracy on these questions. Many questions with indirect references (e.g. ``Did they contact \object{the object they were watching}?'') have an equivalent question with no indirect references (e.g. ``Did they contact \object{a television}?''). The Precision score shows the accuracy of questions with indirect references when the model answered the equivalent question with no indirect references correctly. 

The larger increase in accuracy from Recall to Precision on open answer questions than on binary questions implies that a model better generalizes to open answer questions with indirect references. For the questions for which the model answered the direct version correctly, HME performed better than the other models on binary questions and HCRN performed better on open-ended questions.

\input{tables/full_indirect_refs}
\input{tables/compo_steps}

\textbf{Compositional steps} (Table~\ref{tab:compo_steps}):
Models struggle to generalize to questions with more compositional steps when they train on questions with fewer compositional steps. HME performs the best overall and on binary questions, but HME performs better on open answer questions.

\input{tables/full_overall_results}

\input{tables/reasoning}

\subsection{Scene graph augmentation details} \label{scene_graph_supp}

Action Genome’s spatio-temporal scene graphs~\cite{ji2020action} annotate five sampled frames from each Charades action~\cite{sigurdsson2016hollywood}. Each frame contains object annotations with lists of the contact, spatial, and attention relationships between the subject and the object~\cite{ji2020action}.
We generate questions and answers based on these spatio-temporal scene graphs. Inaccurate or incomplete scene graph data can lead to uninformative and incorrect question generation. Since the scene graph annotations in Action Genome are often noisy, inconsistent, and sparse, we augment them using the following techniques to minimize errors:

\noindent\textbf{Duplication:} When Action Genome contains multiple annotations for the same object, for example if both \object{food} and \object{sandwich} refer to the same object, questions become artificially hard to answer. A person or model who identifies the correct answer to the question ``What were they \relationship{eating}?'' would have $50\%$ chance of answering the question incorrectly if the object is annotated twice.
	
We first tried replacing duplicate references with the hypernyms using WordNet~\cite{miller1995wordnet}, in this case replacing \object{food} with \object{sandwich}. However, the Amazon Mechanical Turk annotators who recorded the Charades videos in their homes used many different objects, including items that do not appear to be \object{sandwiches} to the human eye, to represent \object{sandwiches}. Therefore, merging duplicate annotations into the more specific annotation of \object{sandwich} still lead to inaccuracies. To reduce this sort of error, we resorted to use hypernyms instead of hyponyms. For example, we replace annotations of \object{sandwich} and \object{groceries} with \object{food} throughout the entire dataset. 
	
Sometimes an object is annotated with with multiple references that are not synonymous (e.g~\object{blanket} and \object{clothes}) and therefore can not be replaced throughout the entire dataset. We identify all such objects by identifying objects of different categories with an intersection over union of $\geq0.5$. We merged such object pairs by manually re-annotating each instance. 

We similarly merge action annotations from Charades to remove duplicate action references. We also replace vague actions, like \action{eating something} with their most specific counterparts, like \action{eating some food}.
	
\noindent\textbf{Inconsistency:} There are also inconsistencies in annotations for the spatial relationships \relationship{beneath} and \relationship{above}. We define a person as \relationship{beneath} an object if the object was at head level or above, and \relationship{above} an object otherwise. However, the Action Genome annotations do not use a consistent rule between or even within object classes. We go through \relationship{beneath} and \relationship{above} relationships for every object, remove annotations for objects that have $\leq95\%$ intra-class consistency, and flip annotations when necessary to make them consistent with our definition.

\noindent\textbf{Sparsity: } Action Genome annotations are also sparse. Since Action Genome annotates $5$ frames per Charades action, objects and relationships from one action are not always annotated in the frames sampled from other co-occuring actions. We propagate annotations to surrounding frames using simple heuristic rules. Since spatial relationships do not have entailments, we do not ask questions using spatial relationships for videos with sparse annotations. We consider the 30\% of videos in which fewer than 60\% of object annotations had spatial relationships to be sparse.

\noindent\textbf{Entailments:} Sometimes, Action Genome annotations do not always include all occurring relationships, leading to incorrect and uninformative questions like Q: ``Were they \relationship{touching} \object{the object they were carrying}?'' A: ``No.'' To address this problem, we curate a list of relationship entailments; for instance, if someone is \relationship{carrying} something, they are also \relationship{holding} and \relationship{touching} it. Actions also entail particular relationships. For instance, \action{snuggling with a pillow} entails that someone is \relationship{snuggling} (a verb) with a \object{pillow} (an object).
Therefore, we create a new class of ``verb'' relationships from the Charades action annotations.

\noindent\textbf{Uncertainty in action localization:} Since the exact time an action begins and ends is often ambiguous, actions that actually occur in sequence are often incorrectly annotated with some overlap, resulting in nonsensical questions. We curate heuristics of common sequences of actions to avoid issues with uncertain action localization. For example, a person must \action{take an object} \temporal{before} \action{holding it} and finish \action{holding it} \temporal{before} \action{putting it somewhere}. If these annotations overlap, we automatically adjust the time stamps such that they do not overlap. Using the same entailments, we assume actions must occur if they are missing. For example, if someone begins \action{holding an object} in the middle of the video, we can assume that just before they were \action{taking an object} from somewhere.


\subsection{Question quality checks}

To create challenging and quality questions, we audit each question before it is added to the benchmark using the following filtering processes:

\noindent\textbf{Rare combinations:} We remove questions with object and relationship pairs that occur less than $10$ times (e.g. ``Were they \relationship{twisting} the \object{doorway}?''). 

\noindent\textbf{Blacklisted object-relationship pairs:} Questions also cannot involve object-relationship pairs from a list we manually curated of pairs that are likely to occur in the video but not be annotated in the spatio-temporal scene graph (e.g. ``Were they \relationship{above} the \object{floor}?'' or ``Were they \relationship{wearing} \object{clothes}?''). 

\noindent\textbf{Answer in the question:} If questions indicate the answer (e.g. ``What were they \relationship{twisting} \temporal{while} \action{twisting a blanket?}''), we remove them. 

\noindent\textbf{Realistic decoy questions:} For questions that ask if some action occurred, we create realistic decoys by only asking about actions in which the action's relevant object or verb exists in the video. For example, if a video includes the action \action{opening a door}, we include questions that have the same verb (e.g. ``Did they \action{open a refrigerator}?'') or the same object (e.g. ``Did they \action{fix a door}?''). However, we do not include questions that ask about actions that do not overlap with any objects or verbs in the video (e.g. ``Did they \action{wash a window}?''). 

\noindent\textbf{Confusing relationships:} Action Genome contains attention relationships (\relationship{looking at} and \relationship{not looking at}) and also sometimes annotates the lack of a contact relationship (\relationship{not contacting}). Our human evaluations uncovered that questions with these relationships, such as ``Were they \relationship{looking at} \object{the object they were behind} \temporal{before} \action{walking through the doorway}?''  are hard to answer correctly.
Therefore we do not ask questions with these relationships in our benchmark. 

\noindent\textbf{Similar objects:} We avoid asking about similar pairs of words like \object{door} and \object{doorway} within the same question. 

\noindent\textbf{Multiple possible answer:} For each question, we check that the constraints of the question only lead to one possible answer. For example, if they are holding multiple things we cannot ask ``What were they \relationship{holding}?'' 

\noindent\textbf{Grammatical Correctness:} 
 We ensure grammatical correctness by specifying in each template the necessary tense for any relationship or action and the relevant articles for each object.

\input{tables/program_modules}

\input{tables/structurals}

\input{tables/semantics}

\noindent\textbf{Only one possible answer:} We ignore questions where there is only one possible answer in the entire dataset (e.g. since the verb \relationship{turning off} is only ever associated with the object \object{light}, we do not allow the question ``What were they \relationship{turning off}?'').

\noindent\textbf{Action localization checks: } We only allow an action to be referenced as the ``longest'' or ``shortest'' action if it is $7$ seconds longer or shorter than all other actions. Similarly, we only allow questions comparing the length of two actions if they have more than a $7$ second difference between them. Since localizing the beginning and end of actions is noisy, the $7$ second buffer ensures that actions have a large enough difference in length reduce incorrect questions.


\subsection{Templates}
\begin{figure*}[t]
     \centering
     \includegraphics[width=0.85\linewidth]{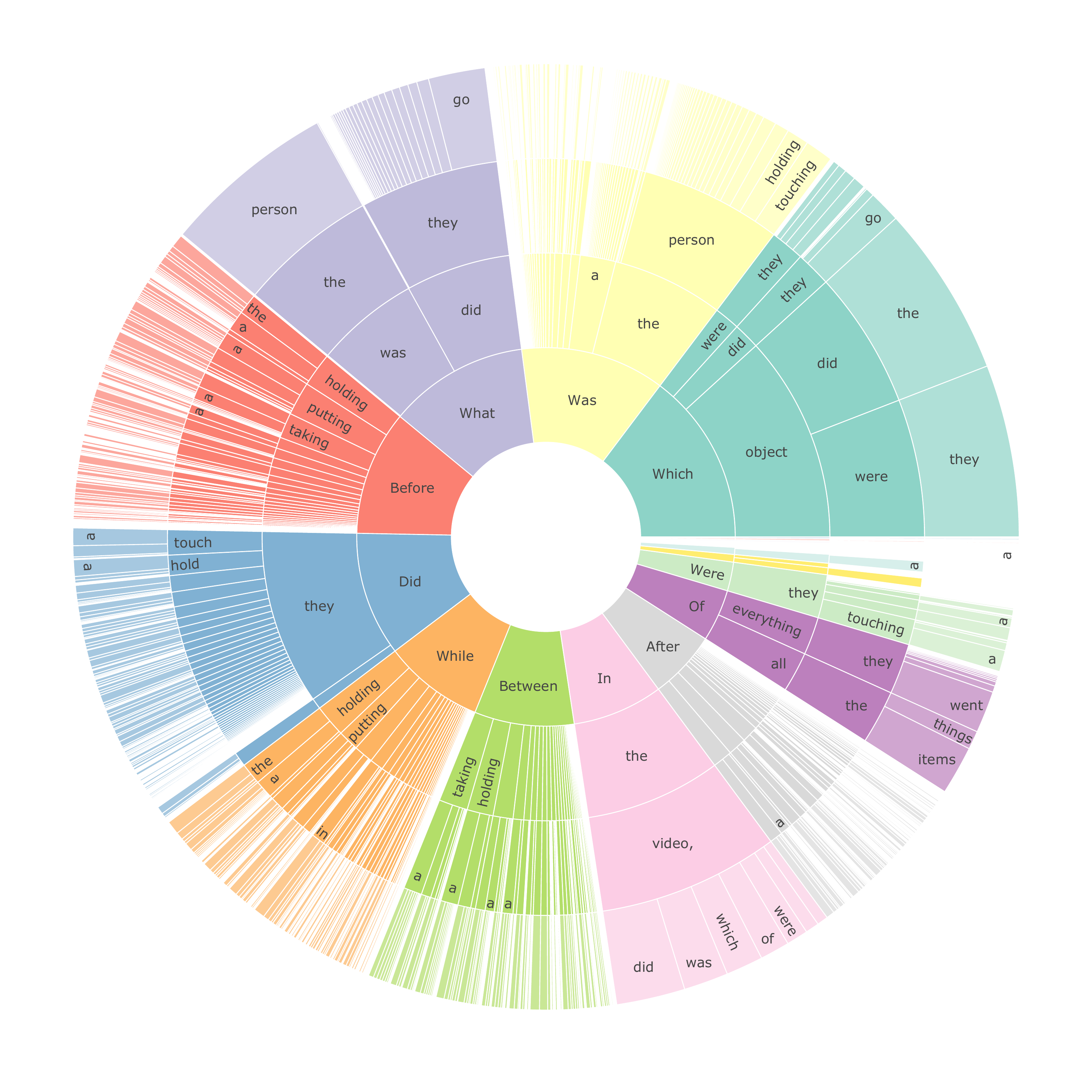}
     \caption{Although the questions in AGQA are generated from just $28$ templates, they are linguistically diverse. There are $3.9$ million total questions in the balanced benchmark and $2.39$ million uniquely worded questions. Above are the first four words of all questions in the balanced benchmark, beginning from the center ring and moving outwards.}
     \label{fig:sunburst}
 \end{figure*}

Our $28$ templates generate AGQA's question-answer pairs (Table~\ref{tab:templates}). Each template has multiple natural language options that can be filled with scene graph information to create a diverse set of questions (Figure~\ref{fig:sunburst}). The templates are also each associated with a program that automatically generates the answer to questions using the spatio-temporal scene graph. These programs are composed of a discrete set of reasoning steps that can be combined in numerous ways to answer a wide variety of questions (Table~\ref{tab:program_modules}). 
Each question has several group labels describing the question's required reasoning skills (Table~\ref{tab:reasoning}), semantic class (Table~\ref{tab:semantics}), and structural category (Table~\ref{tab:structurals}).

 \begin{figure*}[t]
     \centering
     \includegraphics[width=0.75\linewidth]{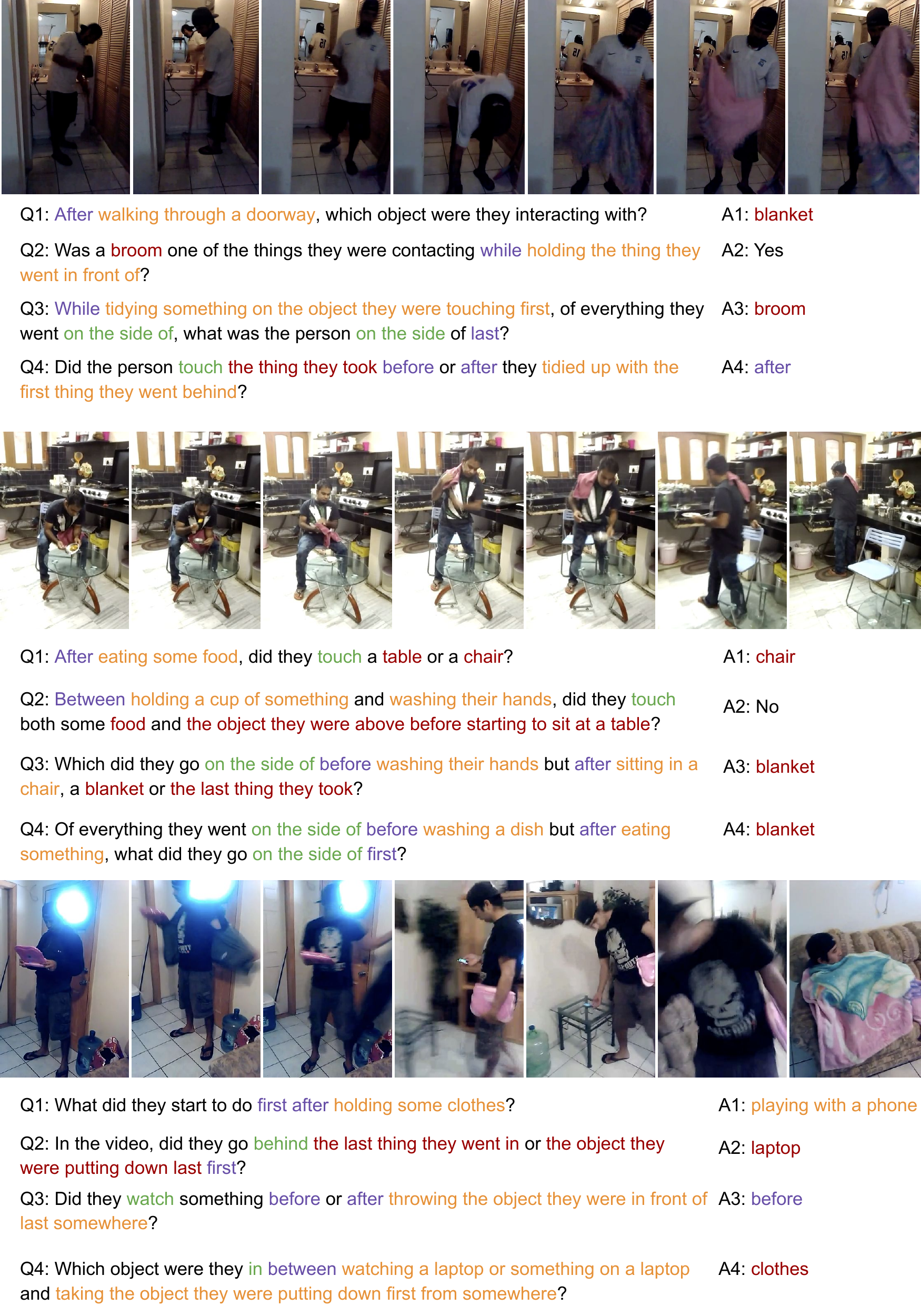}
     \caption{Examples of Questions in AGQA.}
     \label{fig:ex1}
 \end{figure*}

 \begin{figure*}[t]
     \centering
     \includegraphics[width=0.75\linewidth]{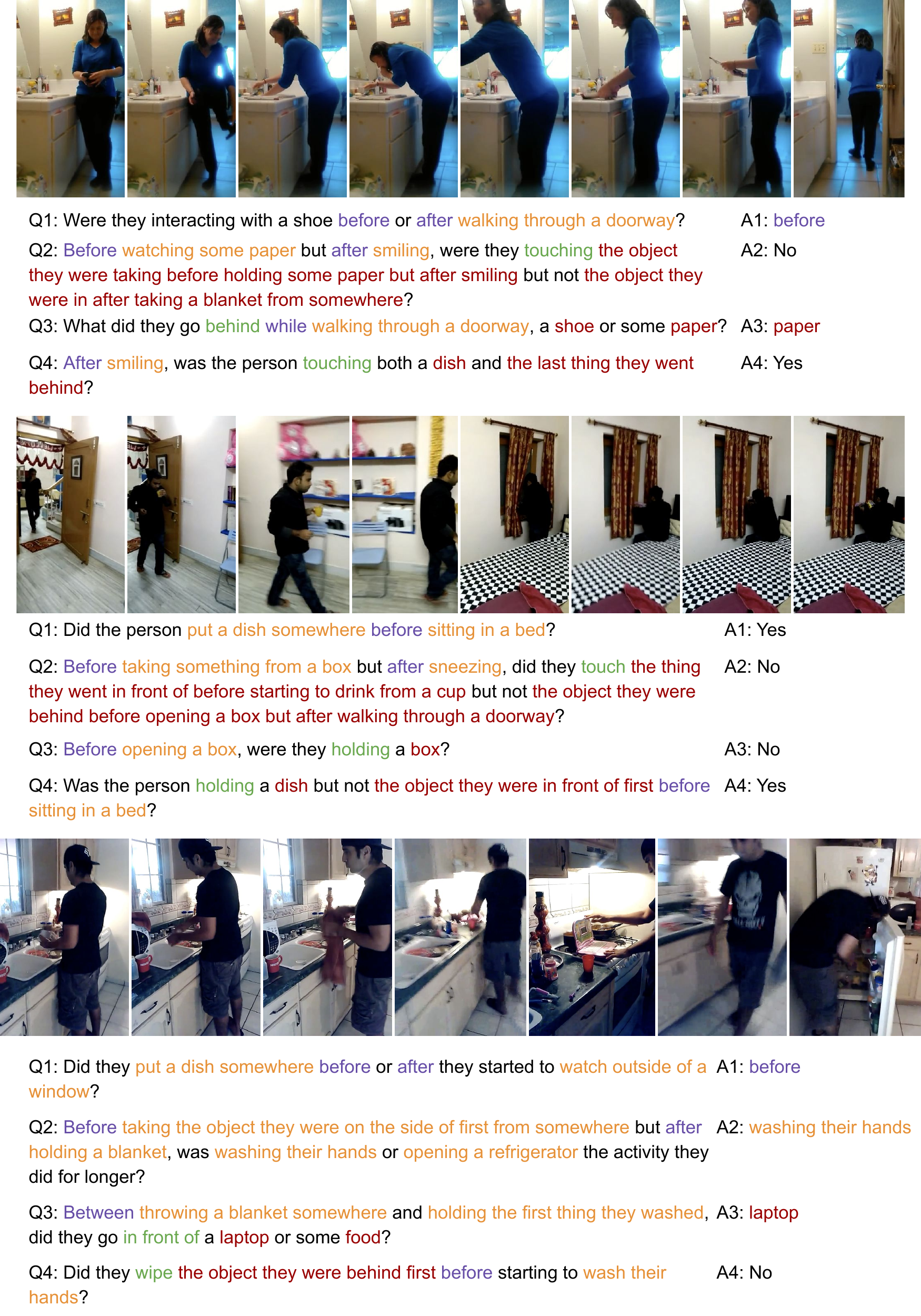}
     \caption{Examples of Questions in AGQA.}
     \label{fig:ex2}
 \end{figure*}

\input{tables/templates}


\subsection{Balancing} \label{balancing_supp}

\input{tables/balancing_algorithm_answers}

\input{tables/balancing_algorithm_structurals}

 \begin{figure*}[t]
     \centering
     \includegraphics[width=0.9\linewidth]{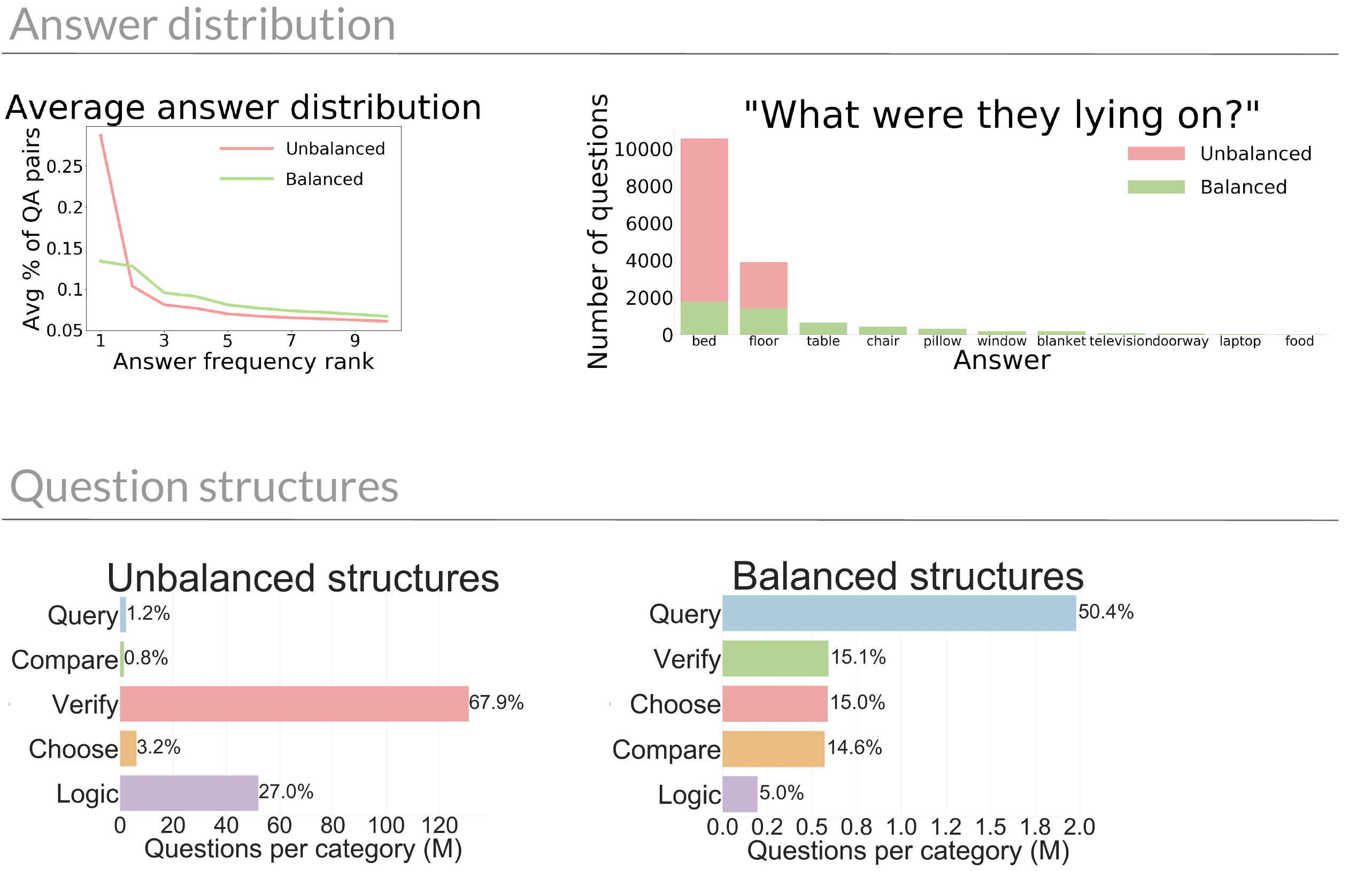}
     \caption{\textbf{Top Row: } The first round of balancing smooths the answer distributions of each question category. The top left figure shows the percentage of questions in the top $10$ frequency ranks of each open answer category. 
     The top right figure shows the answer distribution for all questions with the base ``What were they \relationship{lying on}?'' Before balancing, $64\%$ of all answers were \object{bed}. After balancing, $34\%$ of all answers were \object{bed}. \textbf{Bottom Row:} The second round of balancing deletes questions to change the distribution of question structures. Since query questions are more varied and more difficult, we make them the largest portion of the benchmark.}
     \label{fig:balancing}
 \end{figure*}

Our balancing process occurs in two rounds (Figure~\ref{fig:balancing}). First, we smooth the answer distributions for open answer questions and make each possible answer of binary questions equally likely (Algorithm~\ref{alg:answer_balancing}). Then, we change the proportion of questions of each structure type to create a more diverse and challenging benchmark (Algorithm~\ref{alg:struct_balancing}).

Across all balancing steps in this process, the algorithm deletes questions from a specified question category. For example, the exists-\object{paper} category includes all questions asking if the person contacts some paper in the video. If at any point a category only has one possible answer, all questions from that category are deleted. Within a category with multiple possible answers, we split questions further by the effect of their temporal localization phrase. A temporal localization phrase combines $<$\temporal{time}$>$ and $<$\action{action}$>$ phrases to focus the reasoning process on a segment of the video (e.g ``\temporal{before} \action{washing a dish}''). Many questions with temporal localization phrases (e.g. ``What did they \relationship{put down} \temporal{last before} \action{washing a dish}?'') correspond to an identical question without the temporal localization phrase (e.g. ``What did they \relationship{put down} \temporal{last}?''). Sometimes adding the temporal localization phrase changes the answer of the question. In this example, the answer would change if they \relationship{put down} something \temporal{after} \action{washing a dish}. In other cases, adding the temporal localization phrase does not change the answer. Although many more of the generated questions are in the latter category, where the temporal localization does not change the answer, questions where the temporal localization does change the answer are more difficult. We delete questions such that the number of instances in which a temporal localization phrase changes the answer is close to the number of times it does not.

\noindent\textbf{Answer distributions: } Binary answer distributions are first split into very specific content categories. For example, questions that ask ``Did they \relationship{lie} on a \object{bed} or the \object{floor} \temporal{first}?'' have the content category \temporal{first}-\relationship{lie}-\object{bed}-\object{floor}, with two answers \object{bed} and \object{floor}. We delete questions from the answer that is more frequent until both answers occur an equal number of times. We balance each individual content category, rather than binary questions overall, to reduce a model's ability to guess the right answer based on the question. 

We then smooth answer distributions for open answer questions such as ``What were they \relationship{holding}?'' We define a proportion $b$ to represent the proportion of the ``head,'' or the side of a specified index in the ordered distribution with the more frequent answers, to the ``tail,'' or side of the same specified index in the distribution with the less frequent answers. To avoid errors from very infrequent answers, we ignore the answers that cumulatively represent at most $5\%$ of the question-answers pairs in the distribution. Then, we place the splitting index at the most frequent answer in the distribution and randomly sample to delete questions from the head until either the head to tail ratio is equal to or less than $b$ or deleting any more questions would change the frequency ordering. The splitting index moves down the distribution, and we delete more questions each round. This process smooths the distribution enough such that it reaches a condition that no more than $30\%$ of question-answer pairs have as answers the most frequent $20\%$ of answers types. We defined this condition after experimenting with different parameters to empirically evaluate which condition created the smoothest answer distribution on a wide variety of distribution shapes without deleting more than $90\%$ of the questions. If the splitting index reaches the tail and this condition is not met, the process repeats with a lower $b$ proportion.

We first smooth overall reasoning categories, such as ``superlative,'' and then smooth individual content categories, such as \temporal{first}-\relationship{holding} for the questions that ask ``What were they \relationship{holding} \temporal{first}?'' 

At the end of this balancing round, each general and specific question category has balanced answer distributions that create a more challenging benchmark by reducing, though not eliminating, the model's ability to guess the answer of a large number of questions based on just the questions themselves.

\begin{figure}[t]
    \centering
    \includegraphics[width=\linewidth]{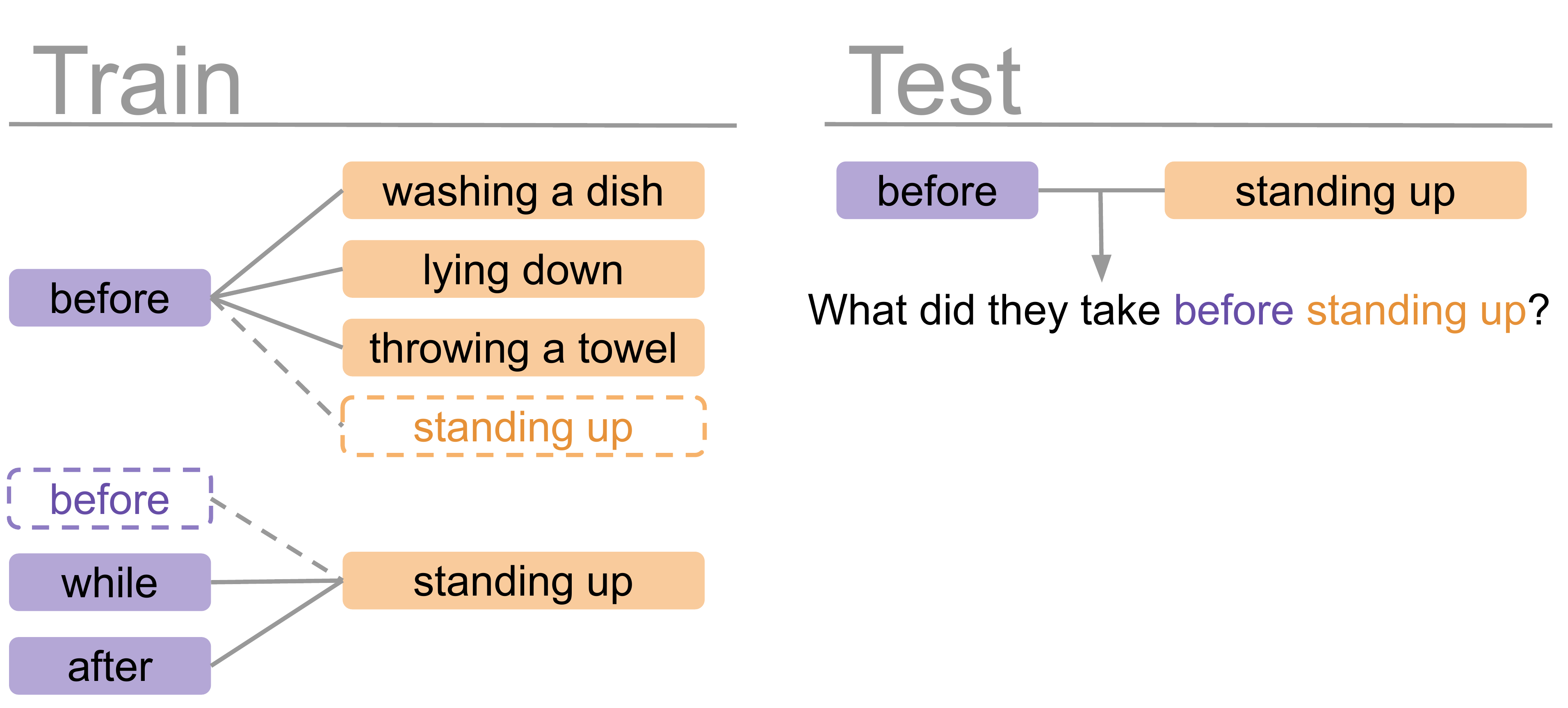}
    \caption{We design the training split for the novel compositions metric by ensuring that certain compositions like \temporal{before} and \action{standing up} occur individually in many questions, but never together in one question. In the test set, we only retain questions where these ideas are combined.}
    \label{fig:novel_compo}
\end{figure}

\noindent\textbf{Question Structures: } After the first round of balancing answer distributions, there are more binary questions than the more difficult open answer questions. We use rejection sampling again to change the distribution of question structure types to increase the proportion of open answered query questions. First, we determine how many questions of each structural type need to be deleted to get close to an ideal structural distribution. However, instead of randomly picking any question of that structural type to delete, we balance the amount of questions to delete to make the distribution of templates and individual question categories as equally spread as possible. Within each distribution questions are deleted such that the original answer distribution holds.

After both rounds of balancing, the benchmark contains a larger percent of open answered and challenging questions, with less skew in the answer distribution.

\begin{figure*}[t]
     \centering
     \includegraphics[width=\linewidth]{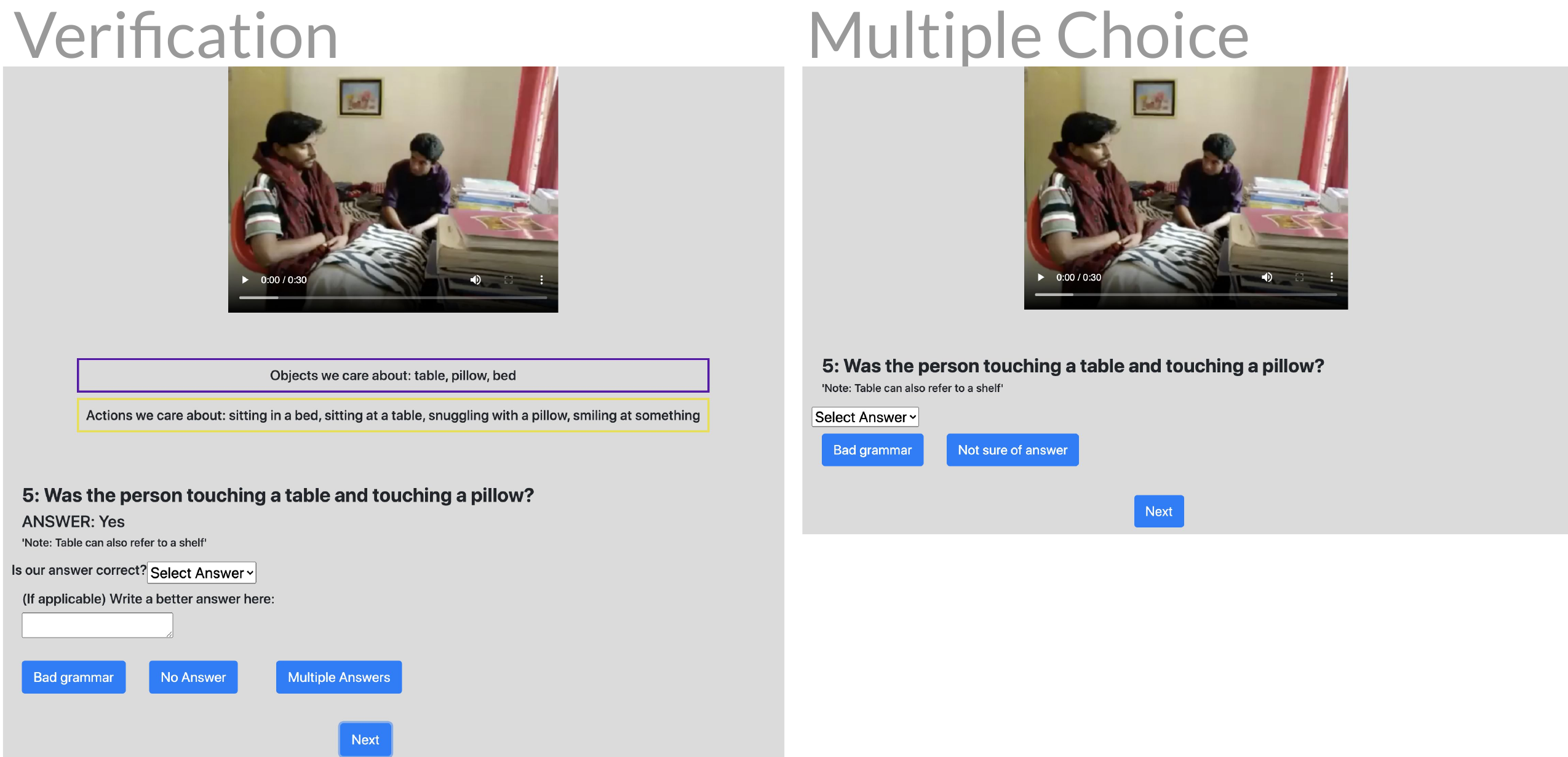}
     \caption{\textbf{Left: } Each annotator watches five videos, each associated with a question and an answer. Annotators indicate if that answer is Correct or Incorrect. \textbf{Right: } The annotators pick the closest answer from a dropdown menu of all activities occurring in the video.}
     \label{fig:amt}
 \end{figure*}


\subsection{Novel compositions}\label{sec:novel_compo}

We explore several types of compositional pairs when constructing the training/test split for the novel compositions metric (Table~\ref{tab:full_novel}).

\noindent\textbf{Sequencing:} To test novel compositions in phrases that localize in time, we select six pairs of \temporal{before}-$<$\action{action}$>$ combinations: \temporal{before}-\action{standing up}, \temporal{before}-\action{walking through a doorway}, \temporal{before}-\action{playing with a phone}, \temporal{before}-\action{opening a laptop}, \temporal{before}-\action{grasping a doorknob}, and \temporal{before}-\action{throwing a broom somewhere}. We selected phrases with a variety in frequency: \action{standing up} occurs very frequently (in $1704$ videos), \action{playing with a phone} somewhat frequently (in $849$ videos), and \action{throwing a broom somewhere} very infrequently (in $29$ videos).
In the test set, there are $55,119$ questions with novel sequencing compositions.

\noindent\textbf{Superlative:} To test novel superlative compositions, we select six compositions of the superlative phrase \temporal{first}-$<$\relationship{relationship}$>$: \temporal{first}-\relationship{behind},
\temporal{first}-\relationship{in},
\temporal{first}-\relationship{leaning on},
\temporal{first}-\relationship{carrying},
\temporal{first}-\relationship{on the side of}, and \temporal{first}-\relationship{holding}. We chose spatial relationships (\relationship{behind}, \relationship{in}, \relationship{on the side of}) and contact relationships (\relationship{leaning on}, \relationship{carrying}, \relationship{holding}).
In the test set, there are $108,003$ questions with such novel superlative compositions.

\noindent\textbf{Duration:} To test novel duration compositions, we select the same six actions used in the sequencing category: \action{standing up}, \action{walking through a doorway}, \action{playing with a phone}, \action{opening a laptop}, \action{grasping a doorknob}, and \action{throwing a broom somewhere}. The test set includes questions that involve the length of these actions. In the test set, there are $10,050$ questions with novel duration compositions.

\noindent\textbf{Object-relationship interaction:} Finally, to test for novel object-relationship compositions, we combine a variety of small and large objects with spatial and contact relationships that each occur frequently.  The pairs we look at are: \object{table}-\relationship{wiping}, \object{dish}-\relationship{wiping}, \object{table}-\relationship{beneath}, \object{dish}-\relationship{beneath}, \object{food}-\relationship{in front of}, \object{paper}-\relationship{carrying}, and \object{chair}-\relationship{leaning on}. Any question that directly asks about this object-relationship pair (e.g Q: ``What were they \relationship{carrying}?'' A: ``\object{paper}''), or that contains this object-relationship pair in an indirect reference (e.g. ``\object{the object they were carrying}''), is removed from the training split and kept in the test split. In the test set, there are $24,005$ questions that contain object-relationship novel compositions. 

\subsection{Human study} \label{amt_task}

We used humans to validate the correctness of AGQA's questions. This section covers the errors annotators found in our benchmark that originate from both our question generation process and those inherited from Action Genome and Charades' human annotation. Some errors enter the question generation process through poor video quality and missing, incorrect, and inconsistent scene graph annotations.
We also found challenges in training annotators with the proper tools and term definitions to effectively annotate question correctness. 

We run two human validation studies, one in which they verify our presented answer, and one in which they choose the correct answer from a dropdown list. We expected to find a performance drop when annotators selected their own answer from the dropdown list as it is a more difficult task to generate one's own answers. By analyzing both studies we can identify what types of questions require higher cognitive effort from annotators and, therefore, lead to larger gaps in performance between the two task formats.

We share these findings in the hope that they synthesize the difficulties in the question-answering task and in inferring visual data from scene graphs. We will conclude with directions for future work on dataset generation to encourage exploration into these problems.

\noindent\textbf{Unclear visual errors in videos: }
Some errors emerge because the objects, actions, and relationships in the video are visually unclear. This uncertainty arises from subtle movements and difficult to see objects. Charades is an visually diverse dataset because crowdworkers filmed the videos. However, this diversity in objects and quality may lead to uncertainty in annotation.

\noindent\textbf{Missing annotations in scene graphs: }
Many of the scene graphs are missing action, object, or relationship annotations. 
In some videos, events occur before the first annotation or after the last annotation, leaving these events at the beginning and end of a video unannotated. Furthermore, some existing objects and relationships were not annotated in Action Genome because they were not a relevant object in a Charades annotation. 
For example, the person in the video may briefly touch a table, but not as a part of any larger action. Therefore, AGQA will answer the question ``Did they \action{touch a table}?'' with ``No,'' even though that object-relationship pair occurs in the video.

Action Genome often had the most salient relationships annotated, but not all relationships. Many missing contact annotations could be added through entailments (e.g. someone \relationship{holding} an object is also \relationship{touching} it). However, not all were recoverable. For example, when \relationship{watching} a \object{phone}, subjects often, but not always, \relationship{touched} it as well. Therefore, we did not add that entailment. Spatial relationships were especially difficult to add through entailments, so we only included questions about spatial relationships for the $70\%$ of videos in which at least $60\%$ of object annotations included a spatial relationship. 

AGQA does not include questions in which the relationship is an answer (e.g. Q: ``What were they doing to the \object{phone}?'' A: ``\relationship{watching}''), because the scene graphs very often missed other relevant relationships that were not annotated. 

Overall, missing annotations caused errors in AGQA's answers in two ways: assuming an existing event did not occur, or assuming there was one answer to a question when there were actually multiple possible answers. We addressed some of these errors through entailments, propagating labels to all annotated frames in an action, creating action priors, and ignoring spatial annotations on videos with sparsity. However, these steps did not fix all errors, and a full overhaul would require large-scale re-annotation efforts.

\noindent\textbf{Incorrect annotations: }
Sometimes, existing annotations were incorrect. For instance, some objects would be annotated as different items than they were in reality. Similarly, one object in the video was sometimes annotated as different objects at different points of the same video (e.g. annotated as a \object{blanket} in some frames and as \object{clothes} in other frames). 

The Charades action annotations were also often incorrect in their start time, end time, and length. Actions that occur in sequence overlapped in their time stamp annotations. For actions of the same family (e.g. \action{taking a pillow} and \action{holding a pillow}), we could infer the sequence and adjust the annotations so they did not overlap. However, this procedure does not work for for actions of different families, so some overlapping annotations remained. Incorrect time stamps also propagate to Action Genome's annotations. Action Genome uniformly sampled $5$ frames from within the action for annotators to annotate. If the action's time stamps were incorrect, these sampled frames may not have been relevant and would have been difficult to annotate.

\noindent\textbf{Incorrect augmentations: }
Incorrect annotations originated mostly from the Action Genome and Charades datasets. However, some were also added by our entailments strategies. For example, when people began \action{holding a dish} in the middle of the video, the annotation \action{taking a dish} was often missing, so we automatically added that annotation. However, in the case when the subject walked into frame in the middle of the video, already \action{holding the dish}, our entailments inserted incorrect actions.

\noindent\textbf{Inconsistent annotations: }
Different annotators appear to have brought different priors on terms and annotation styles. Annotators also use synonymous annotations interchangeably, leading to inconsistent labels (e.g. \action{eating something} and \action{eating some food}).
Annotators also used inconsistent definitions on terms such as \relationship{in front of}, \relationship{behind}, \relationship{above}, \relationship{beneath}, \object{closet}, \relationship{leaning on}, \relationship{snuggling}, \action{sitting down}, and \action{standing up}. These inconsistencies lead to inconsistencies in questions. The question ``Were they \relationship{leaning on} a \object{closet}?'' may have different answers dependent on the annotator's definition of \relationship{leaning on} and \object{closet}. As described in Section~\ref{scene_graph_supp}, we addressed some of the above and beneath inconsistencies by keeping, switching and ignoring them by class, but our mitigation strategies did not solve all inconsistencies.

The annotators were inconsistent along several other fronts as well. Some annotators annotated interactions with the phone that was filming the video, while others ignored those interactions. Some annotators annotated actions performed by animals in the video doing actions like \action{watching out of a window}, while others ignored those actions. Some annotators annotated each individual action separately (e.g. ``\action{eating some food}'' each time the person raised food to their mouth), while others annotated groups of actions (e.g. one ``\action{eating some food}'' annotation for the entire process). We did not include questions about the number of times each action occurred because of these inconsistencies, and we merged overlapping identical annotations.

Annotators also held different priors as to the length of actions that indicate a transition in state (e.g. \action{putting a dish somewhere} and \action{sitting down}). We did not include questions asking about the length of transition verbs to avoid bringing these inconsistencies to our questions.

 \begin{figure}[t]
     \centering
     \includegraphics[width=1\linewidth]{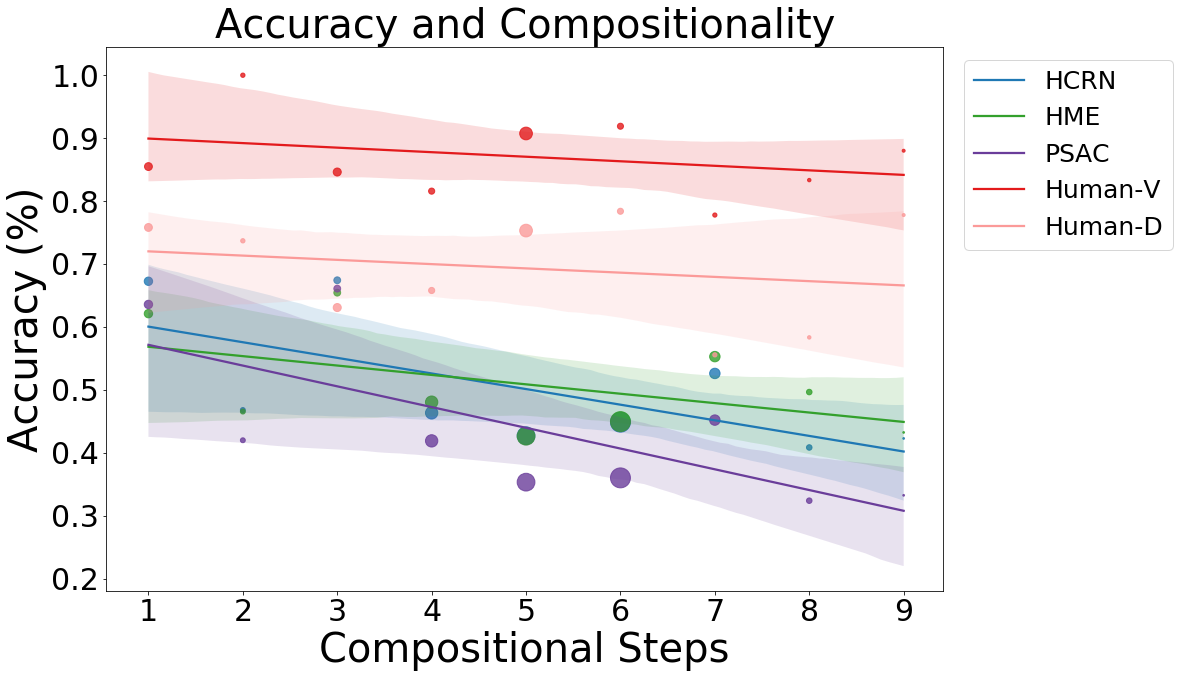}
     \caption{For all three models, we fit a linear regression and find that accuracy is negatively correlated with the number of compositional reasoning steps used to answer the question. Although the $R^2$ scores are relatively weak for all three models: HCRN ($.43$), HME ($.24$), and PSAC ($.51$), the correlation is weaker for both the human verification task (Human-V) and the dropdown task (Human-D) with $R^2$ scores of $.09$ and $.04$ respectively. The size of the dots correlates with the number of questions with each number of steps, with the model's test set size scaled to $1000$x smaller. The shaded area is the $80\%$ confidence interval.
     }
     \label{fig:compo_fig_both}
 \end{figure}

\noindent\textbf{Human and AGQA definition mismatches: }
Similar inconsistencies in term definitions among the annotators who annotated Charades and Action Genome appear in annotators answering AGQA's questions. 
In reducing the effect of synonyms and multiple annotations of the same object causing errors, we combined terms with similar semantic meaning (e.g.~\object{towel} and \object{blanket} are all referred to with the term blanket \object{blanket}). However, the adjusted term may not best describe the item in the annotator's mind. To minimize the effect these errors had on our reported accuracy, we wrote notes next to the question to specify the constraints we used. However, this shift in definition requires extra cognitive effort from the annotator answering the question. 

Annotators also occasionally said the AGQA answer was incorrect, then wrote as a correct answer a term that did not occur in the dataset. Similarly, annotators did not know constraints on the possible object-relationship pairs, so they inferred some pairs that do not exist in AGQA (e.g. \action{watching a pillow}). 

\noindent\textbf{Explaining these errors to our annotators: }
To evaluate AGQA, we designed our human evaluation protocol by minimizing the errors due to incorrect definitions and missing annotations. We designed a qualification task that introduced these different errors to annotators and only allowed them to evaluate AGQA's questions once they passed the qualification. The qualification task provided detailed instructions on our interface. The annotators were given several examples representing different categories of questions and asked to complete the task. If they did not provide the correct answers in the qualification task, we gave explanations for why their given answers were wrong and did not allow them to proceed until they changed the answer to be correct.

\noindent\textbf{Human evaluation tasks: }
To validate the correctness of our question-answer generation process and determine the percent of questions in of our dataset that include these errors, we run human validation tasks on Amazon Mechanical Turk and pay $\$15$ USD per hour. As it is infeasible to verify all $192M$ questions in AGQA, we randomly sampled a subset of questions such that there are at least $50$ questions per reasoning, semantic, and structural category.

Since the videos are filmed in peoples' homes, they often contain objects and actions outside of the benchmark's vocabulary. Therefore, we indicate which objects and actions are relevant. However, when we asked for free form answers to our questions, annotators gave answers outside of the model's vocabulary. Therefore, we tested human accuracy with a verification task. To provide more insight on which questions require high cognitive effort to answer, we also ran a task in which annotators select the answer from a dropdown menu. We develop our interfaces (see Figure~\ref{fig:amt}) using EasyTurk~\cite{krishna2019easyturk}.

For each task, annotators answered one question for each of $5$ videos. 
To improve annotator quality, we ran an qualification task which prevented annotators from proceeding until they placed the correct answer. To ensure annotators answered the questions under the same set of assumptions as AGQA, we added notes next to questions to clarify terms, if necessary. We crowdsourced $3$ instances of each question and counted the majority vote.

\noindent\textbf{Verification Task: }
The verification task showed annotators a video, a question, and a potential answer (Figure~\ref{fig:amt}). They marked the answer as Correct or Incorrect. 
If they marked the answer as incorrect, we asked them to write a better answer in a textbox. To gather more data, we also asked them to select if the question had bad grammar, multiple answers, or no possible answer. Finally, we added question-answer pairs we knew to be incorrect as a gold standard and to introduce variety. Annotators marked as incorrect $80\%$ of the examples we deliberately made incorrect.

\noindent\textbf{Multiple Choice Task: }
The multiple choice task showed annotators a video, a question, and a dropdown list of potential answers selected from the events in the video (Figure~\ref{fig:amt}). We also allowed them to select if the question had bad grammar or if they felt unsure of the answer. We judged a annotator's response as correct if their choice from the dropdown menu matched our answer.

\noindent\textbf{Task design effect on results: }
As the purpose of the human error analysis is to determine which of the questions in AGQA are correct, we used the verification task's results in the main paper. On both tasks human accuracy levels remained consistent as the number of compositional steps increased (Figure~\ref{fig:compo_fig_both}). 
However, across nearly every category, performance decreased when people were asked to select the question from a dropdown menu (Table~\ref{tab:amt_compare}). Performance decreased more for open-ended questions. This decrease could originate from the higher cognitive load it takes for people to generate the answer or from AGQA answers that are correct but ambiguous. The activity recognition category is especially difficult within the dropdown task. It has high cognitive load because there are on average $7.4$ possible answers in the dropdown menu, and the beginning and endpoints of actions may be ambiguous. Both tasks served to illuminate the source of errors in AGQA that we have described.

\input{tables/amt_compare}

\noindent\textbf{Recommendations for future dataset annotation projects: }
AGQA generates questions referring to specific details in the video. This specificity creates challenging questions that inform us about the weaknesses of existing video understanding models. However, our question generation approach relies on the details of the scene graph and a thorough representation that is difficult and expensive to achieve. 

We present several recommendations for annotation practices of scene graph representations of videos that would help address the above errors. First, we suggest that annotators cover the entire video in order to avoid small actions occurring before or after annotations. Second, the time stamps of action annotations should be sequenced in terms of global context to avoid the overlapping of actions that actually occur in sequence. Third, annotators should have explicit definitions of ambiguous concepts; e.g.~spatial relationships like ``\relationship{above}'' should be clearly annotated with respect to the camera or with respect to the subject. Finally, an ideal representation should avoid polysemy, even if that object can be referred to with multiple terms. For example, in Action Genome a \object{sandwich} is often annotated as both a \object{sandwich} and as \object{food}. Even though they refer to the same object in the video, they provided different bounding boxes and appeared on non-identical sets of frames. A representation with one annotation per object that has hierarchical levels of semantic specificity would ameliorate this issue. 

As future work continues to improve the symbolic representation of videos, benchmarks will be better able to measure detailed video understanding.

\subsection{Conclusion}

Despite the challenges outlined in the supplementary materials, our pipeline produced a large balanced dataset of video-question answer pairs that requires complex spatio-temporal reasoning. Our dataset is challenging, as the state of the art models barely improved over models using only linguistic features. We also contribute three new metrics that measure a model's ability to generalize to novel compositions, indirect references, and more compositional steps. Current state of the art models struggle to generalize on all of these tasks. Furthermore, although humans perform similarly on both simple and complex questions, models' performance decreased as question complexity increased. 

Our benchmark can determine the relative strengths and weaknesses of models on different types of reasoning skills and opens avenues to explore new types of models that can more effectively perform compositional reasoning.

%% file: tables/full_novel_comp.tex
\begin{table}
\caption{Results from the novel compositions metric split by binary and open answer questions. Questions with novel compositions in the superlative, sequencing, and object-relationships categories struggle with open-answered questions. B rows and O rows show results for binary and open answer questions respectively.}
\label{tab:full_novel}
\centering
\resizebox{\linewidth}{!}{%
\begin{tabular}{rrrrrr}
\multicolumn{1}{l}{}         & \multicolumn{1}{l}{} & Most Likely & PSAC  & HME            & HCRN           \\ \hline
\multirow{3}{*}{Sequencing}  & B                    & 50.00       & 51.26 & \textbf{56.94} & 51.66          \\
                             & O                    & 9.83        & 23.99 & 31.24          & \textbf{33.19} \\
                             & All                  & 13.67       & 38.35 & \textbf{44.77} & 42.91          \\ \hline
\multirow{3}{*}{Superlative} & B                    & 50.00       & 40.28 & \textbf{51.62} & 40.68          \\
                             & O                    & 11.75       & 9.74  & 14.32          & \textbf{16.15} \\
                             & All                  & 12.60       & 31.97 & \textbf{41.48} & 34.01          \\ \hline
\multirow{3}{*}{Duration}    & B                    & 50.00       & 54.61 & \textbf{60.32} & 56.00          \\
                             & O                    & 19.33       & 25.27 & 38.02          & \textbf{42.94} \\
                             & All                  & 10.96       & 38.65 & 48.19          & \textbf{48.90} \\ \hline
\multirow{3}{*}{Obj-rel}     & B                    & 50.00       & 35.93 & \textbf{44.81} & 37.85          \\
                             & O                    & 66.32       & 2.66  & 0.00           & \textbf{13.82} \\
                             & All                  & 35.63       & 19.12 & 22.17          & \textbf{25.71}
\end{tabular}
}
\end{table}

%% file: tables/full_indirect_refs.tex
\begin{table}[t]
\caption{Results from the indirect references metric split by binary and open answer questions. Generally, open answer questions see a greater increase in accuracy on questions with indirect references when the model can correctly answer the equivalent question without indirect references. B rows and O rows show results for binary and open answer questions respectively. The Precision values for indirect relationship questions are N/A because none of the direct counterpart questions were answered correctly.}
\label{tab:full_indirects}
\centering
\resizebox{\linewidth}{!}{%
\begin{tabular}{rrrrrrrr}
\multicolumn{1}{l}{}          & \multicolumn{1}{l}{} & \multicolumn{2}{c}{PSAC} & \multicolumn{2}{c}{HME}         & \multicolumn{2}{c}{HCRN}        \\
\multicolumn{1}{c}{}          &                      & Precision    & Recall    & Precision      & Recall         & Precision      & Recall         \\ \hline
\multirow{3}{*}{Object}       & B                    & 57.71        & 52.14     & \textbf{70.44} & \textbf{59.62} & 68.87          & 56.01          \\
                              & O                    & 72.91        & 26.19     & 87.26          & 35.97          & \textbf{91.94} & \textbf{37.32} \\
                              & All                  & 64.82        & 38.64     & 79.16          & \textbf{47.32} & \textbf{81.03} & 46.29          \\ \hline
\multirow{3}{*}{Relationship} & B                    & 40.84        & 43.03     & \textbf{48.60}  & \textbf{53.39} & 46.77          & 51.39          \\
                              & O                    & N/A          & 0.98      & N/A            & 0.00           & N/A            & 3.41           \\
                              & All                  & 40.84        & 24.12     & \textbf{48.60}  & 29.39          & 46.77          & \textbf{29.82} \\ \hline
\multirow{3}{*}{Action}       & B                    & 53.32        & 48.25     & \textbf{68.73} & \textbf{59.66} & 63.92          & 54.81          \\
                              & O                    & 75.69        & 23.75     & \textbf{92.47} & 33.57          & 92.18          & \textbf{33.66} \\
                              & All                  & 64.53        & 34.62     & \textbf{81.68} & \textbf{45.15} & 80.22          & 43.05          \\ \hline
\multirow{3}{*}{Temporal}     & B                    & 50.06        & 47.16     & \textbf{64.50}  & \textbf{58.42} & 61.19          & 53.63          \\
                              & O                    & 74.49        & 27.18     & 87.29          & 36.29          & \textbf{92.27} & \textbf{37.23} \\
                              & All                  & 66.48        & 33.15     & 80.71          & \textbf{42.91} & \textbf{83.92} & 42.13         
\end{tabular}
}
\end{table}

%% file: tables/compo_steps.tex
\begin{table}[t]
\centering
\caption{Results from the compositional steps metric split by binary and open answer questions. The models generalize very poorly to questions with more compositional steps.}
\label{tab:compo_steps}
\resizebox{\linewidth}{!}{
\begin{tabular}{rrrrrr}
\multicolumn{1}{l}{} & \multicolumn{1}{l}{} & \multicolumn{1}{c}{Most Likely} & \multicolumn{1}{c}{PSAC} & \multicolumn{1}{c}{HME} & \multicolumn{1}{c}{HCRN} \\ \hline
More                 & B                    & 50.00                           & 35.39                    & \textbf{48.09}          & 42.46                    \\
Compositional        & O                    & 14.51                           & 28.00           & 33.47                   & \textbf{34.81}                    \\
Steps                & All                  & 12.81                           & 31.13                    & \textbf{39.70}          & 38.00                   
\end{tabular}
}
\end{table}

%% file: tables/full_overall_results.tex
\begin{table*}[h]
\caption{Results split by binary and open questions. B rows and O rows show results for binary and open answer questions respectively.}
\label{tab:full_global}
\centering
\begin{tabular}{crrrrrrrr}
\multicolumn{1}{l}{}        & Question Types               & \multicolumn{1}{l}{} & Most Likely & PSAC  & HME            & HCRN (w/o vision) & HCRN           & Human \\ \hline
\multirow{16}{*}{\rotatebox{90}{Reasoning}} & \multirow{3}{*}{obj-rel}     & B                    & 50.00       & 47.91 & \textbf{57.24} & 52.30             & 52.88          & 78.95 \\
                            &                              & O                    & 11.96       & 27.49 & 36.55          & 36.82             & \textbf{37.49} & 90.90 \\
                            &                              & All                  & 8.82        & 34.75 & \textbf{43.91} & 42.33             & 43.00          & 80.65 \\ \cline{3-9} 
                            & rel-action                   & B                    & 50.00       & 56.84 & 57.84          & \textbf{58.06}    & 56.75          & 90.20 \\ \cline{3-9} 
                            & obj-act                      & B                    & 50.00       & 58.33 & 50.00          & 51.67             & \textbf{63.33} & 93.75 \\ \cline{3-9} 
                            & \multirow{3}{*}{superlative} & B                    & 50.00       & 43.35 & \textbf{56.77} & 49.69             & 50.81          & 81.81 \\
                            &                              & O                    & 8.46        & 12.22 & \textbf{18.52} & 18.51             & 18.49          & 80.77 \\
                            &                              & All                  & 10.29       & 30.51 & \textbf{41.10} & 36.83             & 37.48          & 81.25 \\ \cline{3-9} 
                            & \multirow{3}{*}{sequencing}  & B                    & 50.00       & 60.38 & 60.05          & \textbf{62.51}    & 61.77          & 94.73 \\
                            &                              & O                    & 5.57        & 4.60  & 1.29           & 9.91              & \textbf{10.92} & 85.18 \\
                            &                              & All                  & 49.15       & 59.95 & 59.60          & \textbf{62.11}    & 61.28          & 90.77 \\ \cline{3-9} 
                            & exists                       & B                    & 50.00       & 69.94 & 70.01          & 72.12             & \textbf{72.22} & 79.80 \\ \cline{3-9} 
                            & \multirow{3}{*}{duration}    & B                    & 50.00       & 30.49 & 45.25          & \textbf{46.22}    & 46.05          & 91.89 \\
                            &                              & O                    & 5.60        & 3.26  & 6.25           & 10.33             & \textbf{11.68} & 92.31 \\
                            &                              & All                  & 23.70       & 29.75 & 44.19          & \textbf{45.24}    & 45.10          & 92.00 \\ \cline{3-9} 
                            & activity recognition         & O                    & 4.72        & 3.78  & 3.23           & 7.57              & \textbf{11.21} & 78.00 \\ \hline
\multirow{7}{*}{\rotatebox{90}{Semantic}}   & \multirow{3}{*}{object}      & B                    & 50.00       & 45.10 & \textbf{56.18} & 50.00             & 51.13          & 87.39 \\
                            &                              & O                    & 11.85       & 27.27 & 36.33          & 36.59             & \textbf{37.26} & 90.90 \\
                            &                              & All                  & 9.38        & 32.79 & \textbf{42.48} & 40.74             & 41.55          & 87.97 \\ \cline{3-9} 
                            & relationship                 & B                    & 50.00       & 65.51 & 66.10          & \textbf{67.40}    & 66.71          & 83.58 \\ \cline{3-9} 
                            & \multirow{3}{*}{action}      & B                    & 50.00       & 57.91 & 58.87          & \textbf{61.68}    & 61.09          & 90.21 \\
                            &                              & O                    & 4.20        & 3.68  & 3.84           & 8.12              & \textbf{11.31} & 80.95 \\
                            &                              & All                  & 32.91       & 57.91 & 58.12          & \textbf{60.95}    & 60.41          & 86.45 \\ \hline
\multirow{5}{*}{\rotatebox{90}{Structure}}  & query                        & O                    & 11.76       & 27.20 & 36.23          & 36.50             & \textbf{37.18} & 83.53 \\
                            & compare                      & B                    & 50.00       & 56.68 & 58.06          & \textbf{59.65}    & 58.77          & 92.53 \\
                            & choose                       & B                    & 50.00       & 33.41 & \textbf{49.32} & 39.52             & 40.60          & 83.02 \\
                            & logic                        & B                    & 50.00       & 67.48 & 69.75          & 69.47             & \textbf{69.90} & 70.69 \\
                            & verify                       & B                    & 50.00       & 68.34 & 68.40          & 70.94             & \textbf{71.09} & 88.26 \\ \hline
                            & \multirow{3}{*}{Overall}     & B                    & 50.00       & 54.19 & \textbf{59.77} & 57.93             & 58.11          & 86.65 \\
                            &                              & O                    & 11.76       & 27.20 & 36.23          & 36.50             & \textbf{37.18} & 83.53 \\
                            &                              & All                  & 10.35       & 40.40 & \textbf{47.74} & 47.00             & 47.42          & 86.02
\end{tabular}
\end{table*}

%% file: tables/reasoning.tex
\begin{table*}[t]
\caption{A list of the overall reasoning types of questions in AGQA.}
\label{tab:reasoning}
\centering
\resizebox{\linewidth}{!}{%

\begin{tabular}{llllll}
\hline
                     Reasoning type & Templates           & Unbalanced (M)         & Balanced (K)           & Answering question involves                                                   & Example templates                                                                                        \\ \hline
Obj-Rel              & \multicolumn{1}{r}{11} & \multicolumn{1}{r}{81.237}                      & \multicolumn{1}{r}{3014.86}   & A specific interaction with a specific object         & Was the person \textless{}\relationship{relationship}\textgreater \textless{}\object{object}\textgreater{}?                      \\
                     &                        &                           &                           &                                                                & What were they \textless{}\relationship{relationship}\textgreater{}?                                                    \\
                     &                        &                           &                           &                                                                & Were they \textless{}\relationship{relationship}\textgreater \textless{}\object{object}\textgreater \temporal{first}?                       \\\hline
Rel-Action           & \multicolumn{1}{r}{1}  & \multicolumn{1}{r}{0.392} & \multicolumn{1}{r}{206.11}  & A relationship compared to an action  & Were they \textless{}\relationship{relationship}\textgreater something \temporal{before} or \temporal{after} \textless{}\action{action}\textgreater{}? \\\hline
Obj-Act              & \multicolumn{1}{r}{1}  & \multicolumn{1}{r}{0.006}  & \multicolumn{1}{r}{0.48}  & An object in comparison to an action & Where they contacting \textless{}\object{object}\textgreater \temporal{before} or \temporal{after} \textless{}\action{action}\textgreater{}?     \\\hline
Superlative          & \multicolumn{1}{r}{10} & \multicolumn{1}{r}{8.877}  & \multicolumn{1}{r}{961.65}  & An extreme instance of an attribute              & Were they \textless{}\relationship{relationship}\textgreater \textless{}\object{object}\textgreater \temporal{first}?                       \\
                     &                        &                           &                           &                                                                & What was the person doing for the most time?                                                             \\
\hline
Sequencing           & \multicolumn{1}{r}{3}  & \multicolumn{1}{r}{0.927} & \multicolumn{1}{r}{320.39} & The sequence in which two actions occur               & Did they \textless{}\action{action}\textgreater \temporal{before} or \temporal{after} they \textless{}\action{action}?                           \\
                     &                        &                           &                           &                                                                & What did they do \temporal{after} \textless{}\action{action}\textgreater{}?                                                  \\
                     &                        &                           &                           &                                                                & What did they do \temporal{before} \textless{}\action{action}\textgreater{}?                                                 \\\hline
Exists               & \multicolumn{1}{r}{6}  & \multicolumn{1}{r}{176.485} & \multicolumn{1}{r}{590.35} & Verifying if some concept exists                      & Were they \textless{}\relationship{relationship}\textgreater \textless{}\object{object}\textgreater{}?                           \\
                     &                        &                           &                           &                                                                & Did they \textless{}\action{action}\textgreater{}?                                                                \\
                     &                        &                           &                           &                                                                & Did they \textless{}\relationship{relationship}\textgreater something?                                                  \\\hline
Duration comparison             & \multicolumn{1}{r}{6}  & \multicolumn{1}{r}{0.160} & \multicolumn{1}{r}{53.20} & The length of time of actions                         & What did they spend the most amount of time doing?                                                       \\
                     &                        &                           &                           &                                                                & Was \textless{}\action{action}\textgreater something they spent less time doing than \textless{}\action{action}\textgreater{}?           \\
                     &                        &                           &                           &                                                                & Did they \textless{}\action{action}\textgreater or \textless{}\action{action}\textgreater for more time?                   \\\hline
Activity recognition & \multicolumn{1}{r}{2}  & \multicolumn{1}{r}{0.012} & \multicolumn{1}{r}{11.65} & Determining what action occurs                            & What did they do \temporal{after} \textless{}\action{action}\textgreater{}?                                                  \\
                     &                        &                           &                           &                                                                & What did they do \temporal{before} \textless{}\action{action}\textgreater{}?                                                
\end{tabular}

}
\end{table*}

%% file: tables/program_modules.tex
\begin{table*}[]
\caption{Listed are the reasoning steps used to generate an answer from a spatio-temporal scene graph. Items in these inputs and outputs are object, relationship, action, and frame nodes in the spatio-temporal scene graphs.}
\label{tab:program_modules}
\centering
\resizebox{\linewidth}{!}{%
\begin{tabular}{llll}
\hline
Category                      & Reasoning step & Inputs                             & Outputs                                                          \\ \hline
\multirow{6}{*}{Filtering}    & query          & item, attribute type               & attribute                                                        \\
                              & getFrames      & frame, scene graph, before/after   & frames before or after indicated index                           \\
                              & exists         & items, query item                  & true if query item in items, false otherwise                     \\
                              & objectRelation & frame, object, relationship        & true if object relationship exists in the frame, false otherwise \\
                              & chooseOne      &    items, query item1, query item2                                & which of the two items is present in the list                    \\
                              & iterate        & items, function, integer x         & the first x items in data structure to return True in function   \\ \hline
\multirow{3}{*}{Verification} & verify         & boolean                            & ``Yes'' if true, ``No'' if false                                     \\
                              & and            & list of booleans                   & true if all items in list are true, false otherwise              \\
                              & xor            & two booleans                       & true if exactly one boolean is true                              \\ \hline
\multirow{4}{*}{Comparison}   & equals         & item1, item2                       & true if item1 equals item 2 false otherwise                      \\
                              & comparative    & item1, item2, attribute, more/less & item that is more or less in reference to a certain attribute                \\
                              & superlative    & items, attribute, most/least       & the item with the most/least in a certain dimension              \\
                              & difference    & value1, value2       & the difference between the two values              \\\hline
\multirow{4}{*}{Item Sets}    & overlap        & items1, items2                     & true if overlap between items1 and items2, false otherwise       \\
                              & containedIn    & items1, items2                     & true if items1 contained in items2, false otherwise              \\
                              & sort           & items, attribute                   & sorted concepts by attribute                                     \\
\end{tabular}
}
\end{table*}

%% file: tables/structurals.tex
\begin{table*}[t]
\caption{Every question in AGQA is associated with one of these five question structures.}
\label{tab:structurals}
\centering
\resizebox{\linewidth}{!}{%
\begin{tabular}{lrrrll}
\hline
                         & \multicolumn{1}{l}{Templates} & \multicolumn{1}{l}{Unbalanced (M)} & \multicolumn{1}{l}{Balanced (M)} & Description                                        & Example templates                                                                                                        \\ \hline
\multirow{3}{*}{Query}   & \multirow{3}{*}{10}              & \multirow{3}{*}{2.2}                    & \multirow{3}{*}{1.98}               & \multirow{3}{*}{Open ended questions}              & Which object did they \textless{}\relationship{relationship}\textgreater{}?                                                                 \\
                         &                                  &                                       &                                     &                                                    & What did the person do \textless{}\temporal{time}\textgreater \textless{}\action{action}\textgreater{}?                                      \\
                         &                                  &                                       &                                     &                                                    & What did they spend the longest amount of time doing?                                                                    \\ \hline
\multirow{2}{*}{Compare} & \multirow{2}{*}{7}               & \multirow{2}{*}{1.5}                 & \multirow{2}{*}{0.57}               & \multirow{2}{*}{Compare attributes of two options} & Compared to \textless{}\action{action}\textgreater{}, did they \textless{}\action{action}\textgreater for longer?                          \\
                         &                                  &                                       &                                     &                                                    & Did the person contact \textless{}\object{object}\textgreater  \temporal{before} or \temporal{after} \textless{}\action{action}\textgreater{}?                    \\ \hline
\multirow{3}{*}{Choose}  & \multirow{3}{*}{3}               & \multirow{3}{*}{6.1}                 & \multirow{3}{*}{0.59}              & \multirow{3}{*}{Choose between two options}        & Was \textless{}\object{object}\textgreater or \textless{}\object{object}\textgreater the thing they \textless{}\relationship{relationship}\textgreater{}?  \\
                         &                                  &                                       &                                     &                                                    & Did they \textless{}\relationship{relationship}\textgreater \textless{}\object{object}\textgreater or \textless{}\object{object}\textgreater \temporal{first}?       \\
                         &                                  &                                       &                                     &                                                    & Which did they \textless{}\relationship{relationship}\textgreater \temporal{last} \textless{}\object{object}\textgreater or \textless{}\object{object}\textgreater{} \\ \hline
\multirow{4}{*}{Verify}  & \multirow{4}{*}{6}               & \multirow{4}{*}{131.0}                & \multirow{4}{*}{0.59}              & \multirow{4}{*}{Verify if a statement is true}     & Does someone contact \textless{}\object{object}\textgreater{}?                                                                    \\
                         &                                  &                                       &                                     &                                                    & Did they \textless{}\relationship{relationship}\textgreater \textless{}\object{object}\textgreater \temporal{last}?                                         \\
                         &                                  &                                       &                                     &                                                    & Was the person \textless{}\relationship{relationship}\textgreater something?                                                            \\
                         &                                  &                                       &                                     &                                                    & Did they \textless{}\action{action}\textgreater{}?                                                                                \\ \hline
\multirow{2}{*}{Logic}   & \multirow{2}{*}{2}               & \multirow{2}{*}{52.0}                & \multirow{2}{*}{0.19}              & \multirow{2}{*}{Use AND or XOR logical operator}   & Were they \textless{}\relationship{relationship}\textgreater both a \textless{}\object{object}\textgreater and \textless{}\object{object}\textgreater{}?  \\
                         &                                  &                                       &                                     &                                                    & Were they \textless{}\relationship{relationship}\textgreater \textless{}\object{object}\textgreater but not \textless{}\object{object}\textgreater{}?    
\end{tabular}
}
\end{table*}

%% file: tables/semantics.tex
\begin{table*}[t]
\caption{Questions in AGQA are categorized as reasoning primarily about an object, relationship, or action.}
\label{tab:semantics}
\centering
\resizebox{\linewidth}{!}{%
\begin{tabular}{lllll}
\hline
         & Templates           & Unbalanced (M)         & Balanced (M)          & Example templates                                                                                                        \\ \hline
Object   & \multicolumn{1}{r}{11} & \multicolumn{1}{r}{38.1}  & \multicolumn{1}{r}{2.9}  & Were they contacting \textless{}\object{object}\textgreater \temporal{before} or \temporal{after} \textless{}\action{action}\textgreater{}?                      \\
         &                        &                           &                          & Which were they \textless{}\relationship{relationship}\textgreater{}, \textless{}\object{object}\textgreater or \textless{}\object{object}\textgreater{}? \\
         &                        &                           &                          & Was \textless{}\object{object}\textgreater the \temporal{first} thing they were interacting with?                                            \\\hline
Relationship & \multicolumn{1}{r}{5}  & \multicolumn{1}{r}{87.2} & \multicolumn{1}{r}{0.6} & Was the person \textless{}\relationship{relationship}\textgreater \textless{}\object{object}\textgreater{}?                                      \\
         &                        &                           &                          & Did they \textless{}\relationship{relationship}\textgreater something \temporal{before} or \temporal{after} \textless{}\action{action}\textgreater{}?                      \\
         &                        &                           &                          & Was the person \textless{}\relationship{relationship}\textgreater{} something?                                                                  \\ \hline
Action   & \multicolumn{1}{r}{12} & \multicolumn{1}{r}{67.6}    & \multicolumn{1}{r}{0.4}  & Did the person \textless{}\action{action}\textgreater{}?                                                          \\
         &                        &                           &                          & Compared to \textless{}\action{action}\textgreater{}, did they \textless{}\action{action}\textgreater for longer?                          \\
         &                        &                           &                          & Did they \textless{}\action{action}\textgreater \temporal{before} or \temporal{after} \textless{}\action{action}\textgreater{}?                                 
\end{tabular}

}
\end{table*}

%% file: tables/templates.tex
\begin{sidewaystable*}[htbp]
\caption{AGQA's questions come from these 33 templates. Most templates optionally allow phrases that localize within time. }\label{tab:templates}
\centering
\resizebox{\linewidth}{!}{%
\begin{tabular}{lrrlllrl}
\hline
Template                & \multicolumn{1}{l}{Unbalanced (K)} & \multicolumn{1}{l}{Balanced (K)} & Reasoning                         & Structural & Semantic     & \multicolumn{1}{l}{Steps} & Natural language example                                                                                                      \\ \hline
objExists               & 2316.0                            & 99.0                             & exists                            & verify     & object       & 1                         & Did they contact \textless{}\object{object}\textgreater{}?                                                                                   \\
objRelExists            & 14297.9                            & 98.8                             & exists, obj-rel                   & verify     & relationship & 1                         & Was the person \textless{}\relationship{relationship}\textgreater \textless{}\object{object}\textgreater{}?                                                 \\
relExists               & 20465.7                            & 97.6                             & exists                            & verify     & relationship & 1                         & Did they \textless{}\relationship{relationship}\textgreater something?                                                                             \\
actExists               & 66541.1                            & 98.6                             & exists                            & verify     & action       & 1                         & Did they \textless{}\action{action}\textgreater{}?                                                                                           \\
andObjRelExists         & 26010.4                            & 93.4                            & exists, obj-rel                   & logic      & relationship & 3                         & Did they \textless{}\relationship{relationship}\textgreater \textless{}\object{object}\textgreater and \textless{}\object{object}\textgreater{}?                     \\
xorObjRelExists         & 26010.4                            & 97.6                            & exists, obj-rel                   & logic      & relationship & 3                         & Did they \textless{}\relationship{relationship}\textgreater \textless{}\object{object}\textgreater but not \textless{}\object{object}\textgreater{}?                 \\
objWhatGeneral          & 34.2                               & 31.2                             &                                   & query      & object       & 1                         & What did they interact with?                                                                                                        \\
objWhat                 & 1796.6                             & 1574.2                           & obj-rel                           & query      & object       & 2                         & Which object were they \textless{}\relationship{relationship}\textgreater{}?                                                                       \\
objWhatChoose           & 4254.1                            & 194.3                              & obj-rel                           & choose     & object       & 3                         & Which object were they \textless{}\relationship{relationship}\textgreater{}, \textless{}\object{object}\textgreater or \textless{}\object{object}\textgreater{}?     \\
actWhatAfterAll         & 4.1                                & 4.0                              & sequencing, activity-recognition  & query      & action       & 1                         & What did they do \temporal{after} \textless{}\action{action}\textgreater{}?                                                                             \\
actWhatBefore           & 1.4                                & 1.4                              & sequencing, activity-recognition  & query      & action       & 1                         & What did they do \temporal{before} \textless{}\action{action}\textgreater{}?                                                                            \\
objFirst                & 146.2                              & 136.9                            & superlative, obj-rel              & query      & object       & 2                         & Which object were they \textless{}\relationship{relationship}\textgreater \temporal{first}?                                                                   \\
objFirstChoose          & 992.5                             & 197.8                              & superlative, obj-rel              & choose     & object       & 3                         & Which object were they \textless{}\relationship{relationship}\textgreater \temporal{first}, \textless{}\object{object}\textgreater or \textless{}\object{object}\textgreater{}? \\
objFirstVerify          & 1213.4                             & 99.0                            & superlative, obj-rel              & verify     & object       & 3                         & Were they \textless{}\relationship{relationship}\textgreater \textless{}\object{object}\textgreater \temporal{first}?                                                 \\
actFirst                & 5.3                               & 5.0                             & superlative, activity-recognition & query      & action       & 1                         & What were they doing \temporal{first}?                                                                                                         \\
objLast                 & 246.5                             & 223.9                            & superlative, obj-rel              & query      & object       & 2                         & Which object were they \textless{}\relationship{relationship}\textgreater \temporal{last}?                                                                    \\
objLastChoose           & 929.8                             & 196.2                            & superlative, obj-rel              & choose     & object       & 3                         & Which object were they \textless{}\relationship{relationship}\textgreater \temporal{last}, \textless{}\object{object}\textgreater or \textless{}\object{object}\textgreater{}?  \\
objLastVerify           & 5339.0                             & 98.8                             & superlative, obj-rel              & verify     & object       & 3                         & Were they \textless{}\relationship{relationship}\textgreater \textless{}\object{object}\textgreater \temporal{last}?                                                  \\
actLast                 & 1.3                                & 1.2                              & superlative, activity-recognition & query      & action       & 1                         & What were they doing \temporal{last}?                                                                                                          \\
actLengthLongerCompare  & 39.3                                 & 12.6                             & duration-comparison               & compare    & action       & 5                         & Was the person \textless{}\action{action}\textgreater or \textless{}\action{action}\textgreater for longer?                                           \\
actLengthShorterCompare & 39.3                                 & 12.6                             & duration-comparison               & compare    & action       & 5                         & Was the person \textless{}\action{action}\textgreater or \textless{}\action{action}\textgreater for less time?                                        \\
actLengthLongerVerify   & 39.3                                 & 12.6                             & duration-comparison               & compare    & action       & 5                         & Did they \textless{}\action{action}\textgreater for longer than they \textless{}\action{action}\textgreater{}?                                        \\
actLengthShorterVerify  & 39.3                                 & 12.6                             & duration-comparison               & compare    & action       & 5                         & Did they \textless{}\action{action}\textgreater for less time than they \textless{}\action{action}\textgreater{}?                                     \\
actLongest              & 2.3                               & 2.3                            & superlative, duration-comparison  & query      & action       & 1                         & What were they doing for the most amount of time?                                                                                   \\
actShortest             & 0.5                               & 0.5                                & superlative, duration-comparison  & query      & action       & 1                         & What were they doing for the least amount of time?                                                                                  \\
actTime                 & 921.9                              & 315.0                            & sequencing                        & compare    & action       & 5                         & Was the person \textless{}\action{action}\textgreater \temporal{before} or \temporal{after} \textless{}\action{action}\textgreater{}?                                       \\
relTime                 & 391.8                              & 206.1                            & rel-act                           & compare    & relationship & 5                         & Was the person \textless{}\relationship{relationship}\textgreater something \temporal{before} or \temporal{after} \textless{}\action{action}\textgreater{}?                       \\
objTime                 & 6.4                              & 0.5                            & obj-act                           & compare    & object       & 5                         & Did the person contact a \textless{}\object{object}\textgreater \temporal{before} or \temporal{after} \textless{}\action{action}\textgreater{}?                            
\end{tabular}
}
\end{sidewaystable*}

%% file: tables/balancing_algorithm_answers.tex
\SetKwInput{KwInput}{Input}                
\SetKwInput{KwOutput}{Output}              

\begin{algorithm}[b]
\DontPrintSemicolon
  
  \KwInput{$Q$: Unbalanced question-answer pairs}
  \KwOutput{Question-answer pairs with smoothed answer distributions}
  
  \For{reasoning type in reasoning types}{
        \If{reasoning type is binary} {
            $d_{reason}$ = questions to delete to make both answers equally plausible
        }
        \Else{
            $d_{reason}$ = questions to delete so $20\%$ of answers represent at most $30\%$ of all questions
        }
        delete $d_{reason}$ questions from $Q_{reason}$
        
        \For{content category in reasoning type}{
            \If{content category is binary} {
                $d_{content}$ = questions to delete to make both answers equally plausible
            }
            \Else{
                $d_{content}$ = questions to delete so $20\%$ of answers represent at most $30\%$ of questions
            }
            delete $d_{content}$ questions from $Q_{content}$
        }
  }

\caption{Answer distribution smoothing}
\label{alg:answer_balancing}
\end{algorithm}

%% file: tables/balancing_algorithm_structurals.tex
\SetKwInput{KwInput}{Input}                
\SetKwInput{KwOutput}{Output}              

\begin{algorithm}[t]
\DontPrintSemicolon
  
  \KwInput{$Q$: a set of questions with smoothed answer distributions}
  \KwInput{$P$: a map from structural category to a percentage}
  \KwOutput{A set of questions balanced by structural type}
  
  \For{struct in structural categories}{
        $d_{struct}$ = number to delete from $Q_{struct}$ to get $P_{struct}$\\
        $N_{templ}$ = number of templates in struct\\
        split $d_{struct}$ into a $d_{templ}$ for each template such that $Q_{templ} - d_{templ} = Q_{struct} / N_{templ}$ 
        
        \For{templ in structural category}{
            $N_{content}$ = number of content categories in templ\\
        
            split $d_{templ}$ into a $d_{content}$ for each content category such that $Q_{content} - d_{content} = Q_{templ} / N_{content}$ 
            
            \For{content category in template}{
                split $d_{content}$ into a $d_{ans}$ to retain answer distribution.\\
                
                delete $d_{ans}$ questions from $Q_{ans}$
            }
        }
  }
  
\caption{Structural type balancing}
\label{alg:struct_balancing}
\end{algorithm}

%% file: tables/amt_compare.tex
\begin{table}
\caption{We run two tasks on human workers, a verification task in which they verify given answers, and a dropdown task in which they select the answer from a dropdown list. Workers perform better on the verification task, especially on open-ended questions.}
\label{tab:amt_compare}
\centering
\resizebox{\linewidth}{!}{%
\begin{tabular}{crrrr}
\multicolumn{1}{l}{}        & Question Types               & \multicolumn{1}{l}{} & Verification & Dropdown \\ \hline
\multirow{16}{*}{\rotatebox{90}{Reasoning}} & \multirow{3}{*}{obj-rel}     & B                    & 78.95        & 68.42    \\
                            &                              & O                    & 90.90        & 63.64    \\
                            &                              & All                  & 80.65        & 67.74    \\ \cline{2-5} 
                            & rel-action                   & B                    & 90.20        & 78.43    \\ \cline{2-5} 
                            & obj-act                      & B                    & 93.75        & 83.33    \\ \cline{2-5} 
                            & \multirow{3}{*}{superlative} & B                    & 81.81        & 72.73    \\
                            &                              & O                    & 80.77        & 55.77    \\
                            &                              & All                  & 81.25        & 63.54    \\ \cline{2-5} 
                            & \multirow{3}{*}{sequencing}  & B                    & 94.73        & 78.94    \\
                            &                              & O                    & 85.18        & 59.26    \\
                            &                              & All                  & 90.77        & 70.77    \\ \cline{2-5} 
                            & exists                       & B                    & 79.80        & 74.03    \\ \cline{2-5} 
                            & \multirow{3}{*}{duration}    & B                    & 91.89        & 70.27    \\
                            &                              & O                    & 92.31        & 69.23    \\
                            &                              & All                  & 92.00        & 70.00    \\ \cline{2-5} 
                            & activity recognition         & O                    & 78.00        & 54.00    \\ \hline
\multirow{7}{*}{\rotatebox{90}{Semantic}}   & \multirow{3}{*}{object}      & B                    & 87.39        & 74.19    \\
                            &                              & O                    & 90.90        & 60.52    \\
                            &                              & All                  & 87.97        & 72.93    \\ \cline{2-5} 
                            & relationship                 & B                    & 83.58        & 75.37    \\ \cline{2-5} 
                            & \multirow{3}{*}{action}      & B                    & 90.21        & 73.91    \\
                            &                              & O                    & 80.95        & 57.14    \\
                            &                              & All                  & 86.45        & 67.10    \\ \hline
\multirow{5}{*}{\rotatebox{90}{Structure}}  & query                        & O                    & 83.53        & 58.82    \\
                            & compare                      & B                    & 92.53        & 78.16    \\
                            & choose                       & B                    & 83.02        & 66.04    \\
                            & logic                        & B                    & 70.69        & 70.69    \\
                            & verify                       & B                    & 88.26        & 76.93    \\ \hline
                            & \multirow{3}{*}{Overall}     & B                    & 86.65        & 73.85    \\
                            &                              & O                    & 83.53        & 57.93    \\
                            &                              & All                  & 86.02        & 71.56   
\end{tabular}
}
\end{table}